\documentclass[11pt, a4paper, logo, copyright, nonumbering]{antgroup}
\usepackage[square, numbers]{natbib}
\usepackage{dblfloatfix}
\usepackage{ulem}
\usepackage{caption}
\usepackage{dramatist}
\usepackage{xspace}
\usepackage{pifont}
\usepackage{multirow}

\usepackage{xltabular}
\usepackage{longtable}
\usepackage{hyperref}
\usepackage{amsfonts}
\usepackage{amsmath}
\usepackage{amssymb}
\usepackage{lineno}
\usepackage{multirow}
\usepackage{adjustbox}

\usepackage[bottom]{footmisc}

\usepackage{CJKutf8}
\usepackage{subfigure}
\usepackage{setspace}

\usepackage{dsfont}
\usepackage{array}
\usepackage{tabularx}
\usepackage{subfigure}
\usepackage{xcolor}
\usepackage{tabularx}
\usepackage{booktabs}
\usepackage{tikz}

\usepackage{lipsum}
\usepackage{multicol}
\usepackage{authblk}
\usepackage{wrapfig}
\usepackage[most,skins,theorems]{tcolorbox}
\usepackage{array}
\usepackage{pifont}
\definecolor{customgray}{RGB}{230,230,230}
\usepackage{graphicx}

\tcbset{
  aibox/.style={
    width=\linewidth,
    top=8pt,
    bottom=4pt,
    colback=gray!10!white,
    colframe=gray!50!black,
    colbacktitle=gray!70!black,
    enhanced,
    center,
    attach boxed title to top left={yshift=-0.1in,xshift=0.15in},
    boxed title style={boxrule=0pt,colframe=white,},
  }
}
\newtcolorbox{AIbox}[2][]{aibox,title=#2,#1}

\makeatletter
\def\@BTrule[#1]{%
  \ifx\longtable\undefined
    \let\@BTswitch\@BTnormal
  \else\ifx\hline\LT@hline
    \nobreak
    \let\@BTswitch\@BLTrule
  \else
     \let\@BTswitch\@BTnormal
  \fi\fi
  \global\@thisrulewidth=#1\relax
  \ifnum\@thisruleclass=\tw@\vskip\@aboverulesep\else
  \ifnum\@lastruleclass=\z@\vskip\@aboverulesep\else
  \ifnum\@lastruleclass=\@ne\vskip\doublerulesep\fi\fi\fi
  \@BTswitch}
\makeatother

\addto\extrasenglish{
}

 {\begin{list}{}%
         {\setlength{\leftmargin}{#1}}%
         \item[]%
 }
 {\end{list}}

\renewcommand{\phi}{\varphi}

\renewcommand{\leq}{\leqslant}
\renewcommand{\geq}{\geqslant}

\renewcommand{\epsilon}{\varepsilon}
\renewcommand{\imath}{\mathrm{i}}

\newlength{\restsubwidth}
\newlength{\restsubheight}
\newlength{\restsubmoreheight}
\newcommand{\rest}[2]{%
        \settowidth{\restsubwidth}{\ensuremath{#2}}
        \settoheight{\restsubheight}{\ensuremath{{}_{#2}}}
        \ensuremath{{#1\hskip 0.5pt}_{\vrule\kern2pt\parbox[b][%
        4pt][b]{\the\restsubwidth}{%
                        \ensuremath{{}_{#2}}}}}
        }

\reportnumber{001} %

\title{
\vspace{-2.5em}
\centering Open-AoE: An Open Egocentric Manipulation Dataset and Toolchain for Embodied Learning
\vspace{-1em} 
}

\newcommand{\ghlink}{https://github.com/ant-research/Open-AoE}
\newcommand{\hflink}{https://huggingface.co/datasets/inclusionAI/OpenAoE-2000h}
\newcommand{\mslink}{https://www.modelscope.cn/datasets/inclusionAI/OpenAoE-2000h}

\DeclareRobustCommand{\platformicon}[1]{%
  \raisebox{-0.15em}{%
    \includegraphics[height=0.95em,keepaspectratio]{#1}%
  }%
}

\author{%
    \vspace{-0.5em}
  Ant Digital Technologies, Ant Group\\
  {\normalfont\small
    \href{\ghlink}{%
      \platformicon{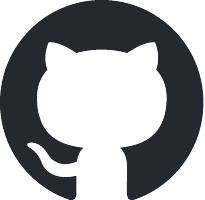}\,
      \texttt{GitHub}%
    }
    \quad$\vert$\quad
    \href{\hflink}{%
      \platformicon{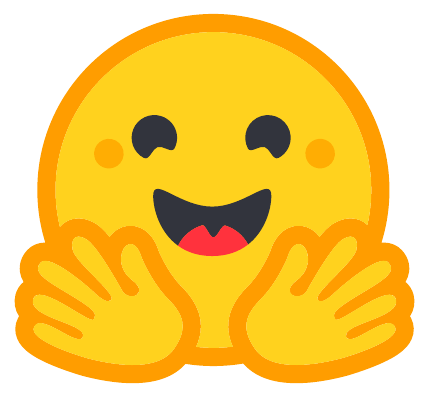}\,
      \texttt{Hugging Face}%
    }
    \quad$\vert$\quad
    \href{\mslink}{%
      \platformicon{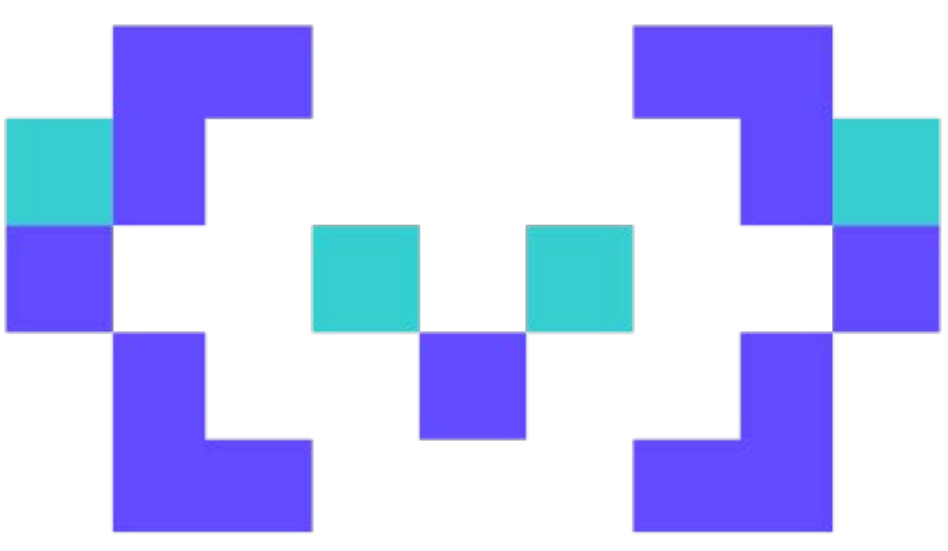}\,
      \texttt{ModelScope}%
    }
  }%
  \vspace{-2em}
}

\begin{abstract}
Egocentric videos of human manipulation provide scalable supervision for embodied intelligence, yet existing resources rarely combine low-cost continuous capture, manipulation-level structured annotations, and reusable tools for robot learning. We present \textbf{Open-AoE}, an open, community-oriented egocentric manipulation dataset and toolchain spanning the full pipeline from smartphone capture to model training. Its first release contains approximately 2,000 hours of manipulation video collected in natural environments by 500+ contributors using 400+ smartphones. The dataset provides text annotations, MANO-based hand poses, camera trajectories, and temporally localized atomic actions. Open-AoE further includes a data processing pipeline that transforms raw recordings into structured samples through temporal action segmentation, semantic annotation, hand reconstruction, and camera trajectory reconstruction. 
Meanwhile, we provide a separate downstream toolchain supports visualization, cross-embodiment retargeting, model-specific data conversion, and training recipes for VLA policies, WAMs, and World Models. 
By integrating scalable capture, structured processing, and downstream adaptation, Open-AoE reduces the barriers to both data contribution and reuse, providing practical open infrastructure for embodied model training, human-to-robot transfer, and world modeling.
\end{abstract}

\begin{document}

\maketitle

\begin{figure*}[htbp]
\centering
\includegraphics[width=0.99\textwidth]{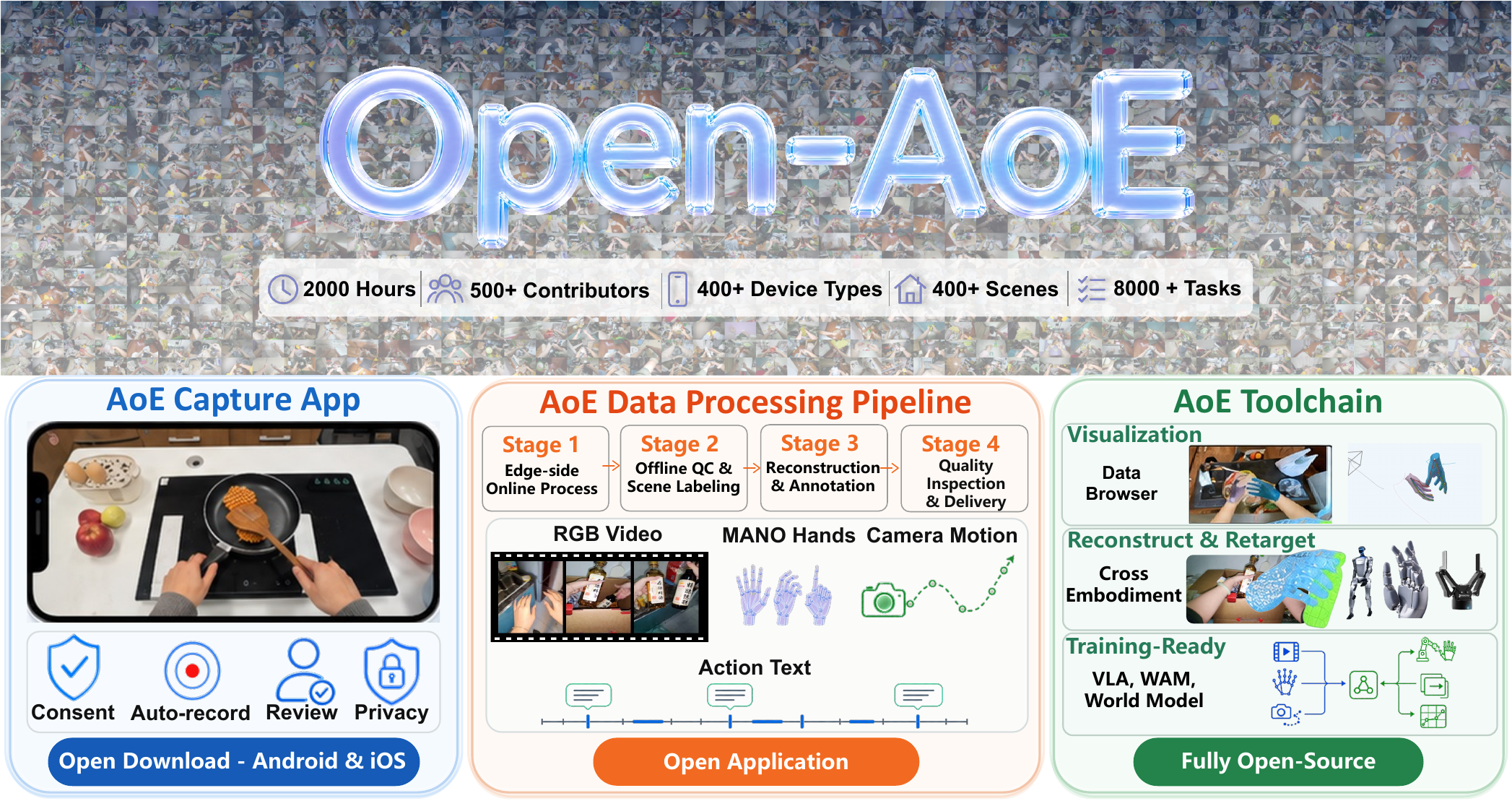}
\caption{\textbf{Overview of Open-AoE.}
Open-AoE provides independent modules for smartphone capture, structured data processing, and downstream robot learning, supporting 2,000 hours of data across 500+ contributors, 400+ device types, 400+ scenes, and 8,000+ tasks.}
\label{figure:fig1}
\end{figure*}
\vspace{-0.5em}


\section{Introduction}

\begin{table}[t]
\centering
\caption{
\textbf{Comparison of egocentric manipulation datasets.} Capture devices are reported as device models or camera types only when explicitly documented by the source; aggregated corpora without a verifiable model inventory are marked Mixed / N/A. Open-AoE combines large-scale consumer-smartphone collection with complete manipulation annotations and a data-to-model toolchain.}
\label{tab:dataset_comparison}
\resizebox{\textwidth}{!}{
\begin{tabular}{lccccccccc}
\toprule
Dataset & Hours & Tasks$^{\dagger}$ & Hand Pose & Language & Camera Traj. & Participants & Capture Devices & Retarget & Training \\
\midrule
Ego4D~\cite{grauman2022ego4d} & 3,670 & N/A & \ding{55} & \ding{51} & \ding{55} & 923$^{\ast}$ & 7 cameras & \ding{55} & \ding{55} \\
EPIC-KITCHENS~\cite{damen2022rescaling} & 100 & N/A & \ding{55} & \ding{51} & \ding{55} & 37 & 2 models & \ding{55} & \ding{55} \\
EgoDex~\cite{egodex2025} & 829 & 194 & \ding{51}$^{\text{native}}$ & \ding{51} & \ding{51} & N/A & 1 model & \ding{55} & \ding{55} \\
EgoLive~\cite{egolive2026} & 1,680 & 346 & \ding{51} & \ding{51} & \ding{51} & N/A & 1 custom model & \ding{55} & \ding{55} \\
OpenEgo~\cite{openegocentric2025} & 1,107 & 290 & \ding{51}$^{\text{unified}}$ & \ding{51} & Partial & N/A & Mixed / N/A & \ding{55} & \ding{55} \\
EgoScale~\cite{egoscale2026} & 20,854 & N/A & \ding{51} & \ding{51} & \ding{55} & N/A & Mixed / N/A & \ding{55} & \ding{55} \\
\midrule
\textbf{Open-AoE} & \textbf{2,000} & \textbf{8,000+} & \ding{51}$^{\text{MANO}}$ & \ding{51} & \ding{51} & \textbf{500+} & \textbf{400+ models} & \ding{51} & \ding{51} \\
\bottomrule
\end{tabular}
}
\parbox{\textwidth}{\footnotesize
$^{\dagger}$Task counts follow source-specific definitions and are not directly comparable.
$^{\ast}$Ego4D uses the \href{https://ego4d-data.org/}{current project-page} counts (923 participants; seven cameras); its CVPR 2022 paper reports 931 camera wearers. OpenEgo's six denotes source datasets, not participants.}
\end{table}

Embodied foundation models are increasingly constrained by the scale and quality of real-world interaction data~\cite{lindata,droid2024}.
Unlike language models, which can learn from internet-scale text, robot learning requires physical demonstrations that capture how humans perceive scenes, move their hands, contact objects, and complete temporally extended tasks~\cite{droid2024}.
Such data must be not only large and diverse, but also structured in geometry, time, and semantics for manipulation learning~\cite{droid2024}.
Empirical scaling studies further show that diversity across environments and objects can matter more than simply increasing the number of demonstrations within fixed conditions~\cite{lindata}.

Egocentric video provides a natural and scalable modality for this problem~\cite{grauman2022ego4d,damen2022rescaling}.
By recording interaction from the actor's viewpoint, it captures hands, objects, scenes, and action progress in the same visual stream~\cite{grauman2022ego4d,damen2022rescaling}.
Compared with third-person video, egocentric video is closer to the perceptual structure of robot execution~\cite{egodex2025,egoscale2026}.
Compared with robot teleoperation, it can be collected more naturally, continuously, and cheaply in everyday environments~\cite{aoe2026,droid2024}.
Recent datasets such as Ego4D, EPIC-KITCHENS, EgoDex, EgoLive, OpenEgo, and EgoScale have advanced egocentric data along multiple axes, including larger-scale daily activity capture, finer-grained hand pose annotation, richer manipulation scenarios, and stronger alignment with robot learning~\cite{grauman2022ego4d,damen2022rescaling,egodex2025,egolive2026,openegocentric2025,egoscale2026}.

However, a central gap remains.
The community does not simply need more egocentric video.
It needs open data infrastructure that can continuously collect, systematically process, and directly support model training.
High-quality egocentric datasets often rely on specialized head-mounted devices, AR/VR hardware, or controlled capture protocols~\cite{egodex2025,egolive2026}.
Specialized hardware can improve pose and camera sensing, but it narrows accessibility relative to commodity-smartphone collection~\cite{egodex2025,egolive2026,aoe2026}.
In contrast, large-scale passive video is easier to obtain, but usually lacks the hand motion, camera trajectory, action boundaries, and physical structure required for robot training~\cite{egodex2025,openegocentric2025}.
Recent efforts have consequently developed dataset-specific pipelines for hand-pose unification, human-to-robot transfer, and raw-video processing~\cite{openegocentric2025,egoscale2026,aoe2026}. 
However, these pipelines remain fragmented and have yet to form a unified infrastructure. This fragmentation underscores the distinction between releasing videos and releasing data that can be used directly for training.

To address this gap, we introduce \textbf{Open-AoE}, a community-oriented egocentric manipulation dataset and toolchain.
Open-AoE builds on the AoE consumer-smartphone collection framework~\cite{aoe2026} and releases the full pipeline from raw capture to model training.
The first release contains approximately \textbf{2,000 hours} of egocentric human manipulation data collected in natural environments by \textbf{500+ contributors} using \textbf{400+ smartphone models}.
The dataset provides structured signals including videos, text descriptions, MANO hand poses, camera trajectories, and atomic action segments.
For dataset production, Open-AoE releases the cloud-side processing pipeline that converts uploaded smartphone clips into structured samples through quality screening and privacy erasure, temporal segmentation, semantic labeling, camera-trajectory estimation, hand reconstruction, atomic-action annotation, and multi-stage quality inspection.
Separately, the downstream data consumption toolchain starts from the released corpus and supports synchronized visual inspection, 4D hand-object interaction reconstruction, cross-embodiment motion retargeting, and robotized-video generation.
The Training-ready toolchain then maps the aligned video, hand, camera, and language signals into action representations and training interfaces for three model directions: Vision-Language-Action (VLA) policies, World Action Models (WAMs), and World Models.

The goal of Open-AoE is not to release an isolated video collection, but to open a reusable \textbf{capture-process-reconstruct-train} data production loop.
Starting from the same egocentric corpus, researchers can inspect data, reconstruct 4D interactions, replay motions on robots, train vision-language-action models, learn world action models, or study world model learning without rebuilding a private post-processing stack.
By lowering both the barrier to contributing data and the barrier to using data, Open-AoE aims to turn low-cost smartphone-captured human manipulation video into practical open infrastructure for embodied intelligence research.

Our contributions are three-fold:
\begin{itemize}
    \item We release approximately \textbf{2,000 hours} of real egocentric human manipulation data collected with consumer-grade smartphones in natural environments, covering diverse participants, devices, and everyday manipulation scenarios.

    \item We open an end-to-end \textbf{data processing pipeline} that converts raw smartphone videos into structured training samples with hand poses, camera trajectories, action segments, and language descriptions.

    \item We provide a downstream \textbf{embodied learning toolchain} for data visualization, reconstruction, and cross-embodiment retargeting, together with training-ready representations and interfaces for three major model directions: VLA policies, WAMs, and World Models.
\end{itemize}

\section{Related Work}

\subsection{Open Egocentric Data Sources for Embodied Learning}

Open egocentric corpora differ chiefly in the supervision they expose above raw video. EPIC-KITCHENS-100~\cite{damen2022rescaling} and Ego4D~\cite{grauman2022ego4d} provide broad coverage of daily activities, environments, and language. Geometry-rich releases make manipulation more directly observable: HOI4D~\cite{liu2022hoi4d} provides RGB-D sequences with hand and object poses, segmentation, and reconstructed geometry; HOT3D~\cite{banerjee2025hot3d} adds motion-capture ground truth for multi-view hand, object, and camera poses; EgoMimic~\cite{kareer2024egomimic} pairs camera motion and 3D hand tracking for imitation learning; and EgoDex~\cite{egodex2025} scales native hand tracking and language annotation to 829 hours of dexterous tabletop manipulation.

Other releases address normalization, scale, or sensor fidelity. OpenEgo~\cite{openegocentric2025} consolidates 1,107 hours from six public datasets into standardized hand poses and localized action primitives. EgoScale~\cite{egoscale2026} studies dexterous policy scaling with more than 20,000 hours of in-the-wild video, while EgoLive~\cite{egolive2026} contributes 1,680 hours of 60-FPS stereo recordings from service scenarios with camera motion, 3D hands, depth, masks, and sub-task descriptions. Open-AoE~\cite{aoe2026} complements these efforts with a 2,000-hour smartphone-captured release spanning 500+ contributors, 400+ device types, 400+ scenes, and 8,000+ tasks, together with atomic action descriptions, MANO hand motion, camera trajectories, and reusable processing and model-integration tools.

\subsection{Embodied Applications of Egocentric Data}

\textbf{Reconstruction and cross-embodiment transfer.}
Egocentric video becomes actionable when implicit motion is converted into geometric or control-oriented supervision. HaWoR~\cite{hawor2025} recovers world-space hand motion from monocular video, while EgoInfinity~\cite{egoinfinity2026} reconstructs metric 4D hand-object trajectories and supports robot retargeting. EgoAERO~\cite{niu2026egoaero} estimates contact-consistent trajectories from a single RGB-D demonstration, and EgoEngine~\cite{liu2026egoengine} jointly generates robotized observations and feasible robot actions. These outputs support several downstream uses: reconstructed hand-object assets expose contact and object motion, retargeted trajectories provide robot-facing supervision, and robotized videos preserve scene context while narrowing the visual embodiment gap. Visual adaptation, shared action spaces, active vision, and whole-body retargeting further address viewpoint and morphology differences across embodiments~\cite{phantom2025,luo2026beingh05,yu2025egomi,yang2026zerowbc,shi2026egohumanoid}.

\textbf{Vision-Language-Action policies.}
VLA methods learn action-aware representations directly from human video rather than requiring every example to be replayed by a robot. Being-H0~\cite{luo2025being} tokenizes human motion for vision-language-action pretraining; EgoVLA~\cite{egovla2025} predicts wrist and hand actions before robot adaptation; and VITRA~\cite{vitra2025} derives robot-aligned supervision from real-life activities. EgoScale~\cite{egoscale2026} further shows benefits from scaling diverse egocentric pretraining. The route from human observation to robot policy can therefore combine language-aligned task semantics with embodiment-aware action targets, but it depends on temporally stable segments, geometry, and action interfaces that remain meaningful after embodiment conversion.

\textbf{World Action Models.}
WAMs occupy a middle ground between direct policy learning and unconstrained video generation: they learn action variables together with observation dynamics, including settings where explicit robot actions are unavailable. Latent-action learning can infer compact motion surrogates from video, although visual distractors may require additional supervision~\cite{laom2025}. LaWAM~\cite{chen2026lawam} uses compact latent visual subgoals to reduce pixel-space generation latency, while DreamZero~\cite{dreamzero2026} jointly predicts future video and action and uses the learned dynamics as a zero-shot policy. This route makes temporal alignment between the current observation, the inferred or supplied action, and the resulting visual change especially important.

\textbf{World Models.}
Predictive world models use egocentric video to learn reusable scene and interaction dynamics beyond a single robot policy. DreamDojo~\cite{dreamdojo2026} pretrains a generalist robot world model on 44,000 hours of egocentric video with continuous latent actions; iVideoGPT~\cite{ivideogpt2024} studies scalable interactive video prediction; and Ctrl-World~\cite{ctrlworld2025} introduces controllable generation for robot manipulation. Such models can exploit both weak latent actions and explicit hand, camera, or robot controls, provided that long videos are segmented into coherent transitions and paired with stable conditioning signals. Across reconstruction, retargeting, VLA, WAM, and world-model learning, the shared requirement is an interface between diverse human video and model-consumable supervision. Open-AoE is designed to provide both sides of that interface through a broad smartphone corpus and its associated processing and conversion tools.

\section{Open-AoE Dataset}
\subsection{Data Processing Pipeline}

\begin{center}
    \centering
    \includegraphics[width=\linewidth]{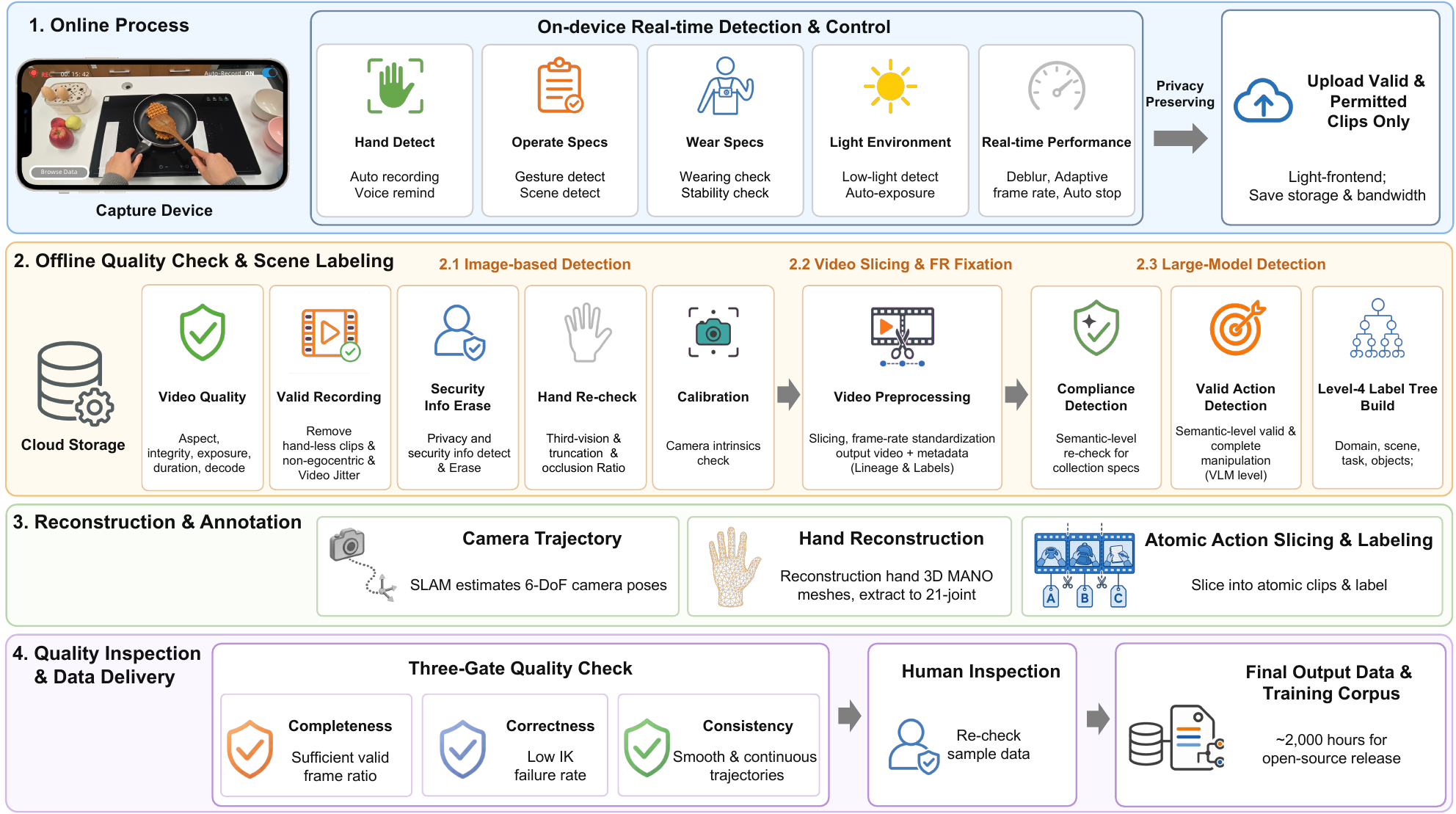}
    \captionof{figure}{\textbf{Open-AoE data processing pipeline.} Starting from edge-side on-device detection and control, raw egocentric videos pass through offline quality checking and scene labeling (image-based detection, video slicing, and large-model detection), followed by pose reconstruction, atomic-action annotation, and three-gate quality inspection to produce an anonymized training corpus. The released subset contains approximately 2,000 hours of egocentric manipulation video captured with consumer smartphones, together with English annotations, MANO hand poses, and camera trajectories.}
    \label{fig:data-pipeline}
\end{center}

As shown in Figure~\ref{fig:data-pipeline}, the pipeline comprises an online capture stage and an offline processing stage.
It proceeds through four stages, namely edge-side online detection, offline quality checking with scene labeling, reconstruction and annotation, and quality inspection with data delivery, and finally produces the anonymized training corpus.

\subsection{Edge-side Online Detection}

On the capture device, lightweight on-device vision models gate recording in real time and govern capture quality, so that only clips containing meaningful hand-object interaction are retained~\cite{aoe2026}.
A hand-visibility check starts recording automatically once hands appear and stops it when they leave the frame, and prompts the user by voice to readjust whenever a hand reaches the image border or becomes truncated, which keeps the manipulating subject fully visible.
Operating and wearing compliance are monitored at the same time. The operating-spec check combines gesture detection with scene detection to flag non-standard actions and non-compliant capture scenes, while the wearing-spec check performs both a wearing check and a stability check to correct improper wearing poses and unstable views; either kind of issue is reported to the user through real-time voice feedback.
The device also adapts to the lighting environment by turning on a constant fill light together with global auto-exposure under dim light, and it sustains real-time performance through a fixed focal length, motion stabilization, deblurring, and an adaptive frame rate, halting recording when storage becomes insufficient or the device overheats.
This lightweight front-end design protects privacy at the source, saves storage and bandwidth, and uploads only valid and permitted clips to the cloud for offline processing.

\subsection{Offline Quality Check \& Scene Labeling}

\subsubsection{Image-based Detection}

The first gate operates on sampled frames without deep models, relying on image-analysis algorithms and lightweight detectors.
For video quality, a bank of rule-based detectors runs in parallel to flag the common defects that make recordings unusable, covering aspect ratio and rotation, file-header and stream integrity, pure-black or over- and under-exposed frames, insufficient duration, and decoding integrity, so that clearly corrupted videos are discarded early.
A valid-recording check then removes clips without visible hands, non-egocentric recordings such as fixed-mount or third-person side views, and segments with severe camera jitter, leaving only valid first-person manipulation.
Security-information erasure detects faces and other sensitive content and masks (pixelates) them in the frames, and it further replaces the personally identifiable information in the metadata, including the collector's name and contact, with anonymized identifiers while rewriting the corresponding fields and directory names; the mapping is stored in full and supports incremental reuse, so the process can be safely re-run and the anonymization reversed when authorized.
A hand re-check then verifies that both hands remain clearly visible by screening for residual third-person views, hand truncation, and an excessive occlusion ratio; a defect is confirmed only when it persists across several consecutive frames, which suppresses single-frame noise.
Calibration finally validates the camera intrinsics for completeness and validity, since reliable calibration is a prerequisite for later pose reconstruction.

\subsubsection{Video Slicing and Frame-Rate Fixation}

After image-based detection, the video is split along contiguous qualified and unqualified spans and its frame rate is fixed, so that a long recording becomes several semantically independent segments (termed \emph{Parts}), with qualified segments that are too short flagged separately.
Each Part stays linked to its source video through a unified process-tag file that records the data lineage from the raw video to each Part and accumulates multi-level labels as the pipeline advances, establishing a data contract that holds throughout the pipeline.

\subsubsection{Large-Model Detection}

Qualified Parts then move to the large-model stage, where the vision-language model Qwen3.7-Plus performs higher-level semantic detection and scene labeling.
Compliance detection re-checks at the semantic level whether a segment follows the collection specifications and identifies violations that are difficult to detect at the image level, while valid-action detection judges whether the action is valid and forms a complete, purposeful manipulation, complementing the earlier image-level screening.
The same model then constructs a four-level label tree under a closed-set design, assigning a representative frame exactly one domain, one scene, and one task from predefined sets together with the salient objects it contains, after which these outputs are standardized into unified identifiers and accumulated over the domain, scene, task, and object dimensions to form a hierarchical label system.

\subsection{Reconstruction \& Annotation}

Each qualified Part enters reconstruction, where camera-trajectory estimation, hand reconstruction, and atomic-action annotation are produced together.
Camera poses are recovered by DROID-W~\cite{droidw2026}, whose robust kernels are re-tuned for handheld and wearable capture to keep the 6-DoF trajectory stable under severe camera shake and torso-motion interference.
Hand reconstruction starts from a two-hand detector trained on large-scale egocentric data collected through AoE, which supplies per-frame bounding boxes, links detections across frames, and fills in missing or occluded hands; building on these boxes, HaWoR~\cite{hawor2025} recovers 3D MANO~\cite{mano2017} meshes that are metric-scale aligned through SLAM and brought into world coordinates, with sliding-window optimization enforcing temporal consistency.
A global bundle adjustment over the HaWoR reconstruction and the dynamic-object masks from DROID-W then optimizes the hand meshes and the camera trajectory jointly in a single world coordinate frame, removing the scale drift and coordinate mismatch that arise across segment-wise reconstructions.
From the recovered meshes, 21-joint hand keypoints are extracted, and each video is finally segmented into semantically coherent atomic clips labeled in English, with a human-in-the-loop review correcting model hallucinations.

\subsection{Quality Inspection \& Data Delivery}

The reconstructed and annotated Parts pass a three-gate quality inspection before archiving, from which about 2{,}000 hours are curated from the full data pool for open-source release.
The completeness gate retains Parts whose hand-reconstruction valid-frame ratio is high enough for reliable pose annotation; the correctness gate retains those whose inverse-kinematics failure rate stays low when retargeting to the 28-DoF joint space, which preserves downstream usability; and the consistency gate retains Parts with smooth, continuous camera trajectories free of abrupt jumps between adjacent frames.
Finally, a randomly sampled subset undergoes human inspection to confirm annotation accuracy and to remove edge cases such as severe occlusion or extreme lighting, yielding the anonymized training corpus that constitutes the final output.

\subsection{Data Distribution Analysis}
\label{sec:data-distribution}

\begin{figure*}[t]
    \centering
    \includegraphics[width=\linewidth]{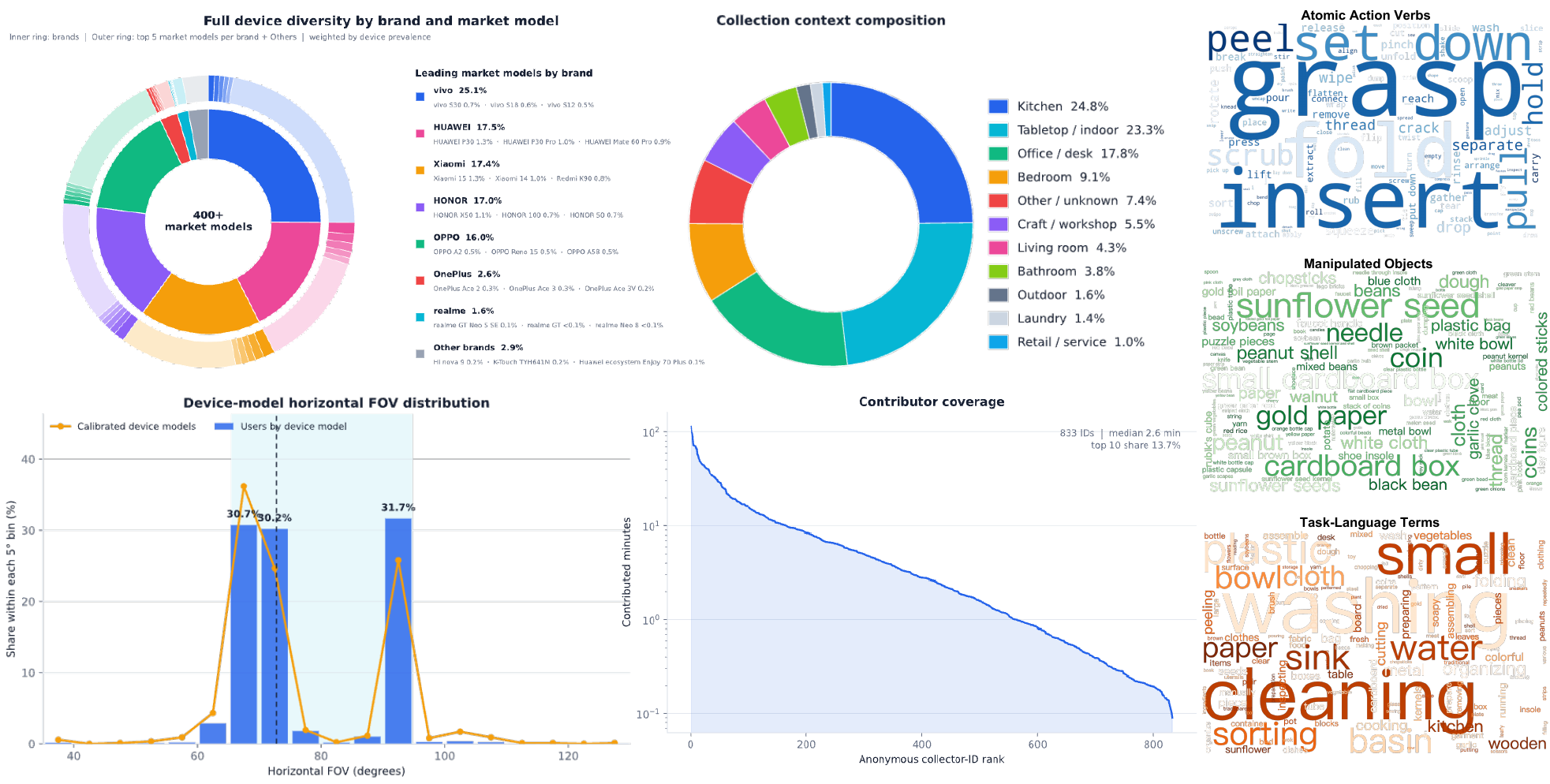}
    \caption{\textbf{Open-AoE collection, sensor, and semantic diversity in a random 100-hour sample.} The nested donut summarizes the consumer-phone brand and market-model composition, and the lower-left panel shows the horizontal-FOV distribution. The center panels report collection-context composition and anonymous collector-ID coverage, while the word clouds summarize atomic-action verbs, manipulated objects, and task-language terms. All panels use the same randomly sampled 100-hour subset; wedge area, bar height, and word size encode relative prevalence or frequency.}
    \label{fig:data-distribution}
\end{figure*}

Figure~\ref{fig:data-distribution} summarizes the diversity observed in a randomly sampled 100-hour subset of the Open-AoE release. All panels use this same sampling scope, including semantic content, collection context, anonymous contributor coverage, consumer-phone models, and horizontal field of view (FOV).
Wedge area, bar height, and word size encode relative prevalence or frequency rather than absolute duration.

\paragraph{Broad scenes, actions, and interaction objects.}
The updated 100-hour audit used in Section~\ref{sec:dataset-analysis} contains \textbf{32{,}407 distinct natural-language action descriptions}, \textbf{175 action verbs}, \textbf{8{,}030 object strings}, and \textbf{135 scene labels}.
As reported in Figure~\ref{fig:temporal_semantic}, the updated temporal annotations achieve \textbf{99.99\% temporal coverage}, with a mean segment duration of \textbf{9.64~s} and an annotation density of \textbf{13.97 segments per minute}.
Temporal coverage is computed from the union of annotated time spans, whereas mean duration and density are computed over annotation records; these quantities should therefore not be multiplied as if the annotations formed a strictly non-overlapping partition. These values supersede the interval and vocabulary counts from an earlier preprocessing snapshot.
These distributions cover reusable manipulation primitives together with a long tail of foods, tools, containers, craft materials, garments, and household objects.
Kitchens, tabletop or indoor settings, offices or desks, and bedrooms account for 24.8\%, 23.3\%, 17.8\%, and 9.1\%, respectively, while the remainder spans living rooms, workshops, bathrooms, outdoor areas, laundry spaces, and retail or service settings.
The joint diversity of scene, action, and interaction object provides a substrate for compositional generalization.

\paragraph{Distributed contributor coverage.}
The same 100-hour sample exhibits a long-tailed distribution across anonymous collector IDs. The median contribution is 2.6 minutes per ID, and the ten largest contributors together account for only 13.7\% of sampled duration.
This long-tailed coverage indicates that the sample is not dominated by a small group of collectors, preserving broader variation in manipulation habits, hand appearance, and recording environments.

\paragraph{A broad consumer-camera domain.}
The random 100-hour sample covers \textbf{400+ consumer-smartphone market models}. The nested donut groups brands in the inner ring and exposes leading models plus the aggregated model tail in the outer ring.
To our knowledge, this is the broadest reported consumer-phone device coverage for an egocentric manipulation dataset.
Restricting device statistics to the same random sample makes the model, semantic, context, and contributor distributions directly comparable under one evaluation scope.
The resulting breadth provides substantial consumer-camera domain diversity within the sampled subset.
Device identity is also a practical proxy for the underlying camera stack, including the lens, sensor, image-signal processor, resolution mode, exposure pipeline, and capture firmware.
Compared with a single-headset collection, Open-AoE therefore introduces naturally occurring camera-domain variation at capture time.

\paragraph{Optical diversity for robust visual transfer.}
Within the random 100-hour sample, the user-weighted horizontal-FOV distribution is multimodal: its three dominant 5-degree bins account for \textbf{30.7\%, 30.2\%, and 31.7\%}, concentrated around $65^{\circ}$--$75^{\circ}$ and $90^{\circ}$--$95^{\circ}$.
The multimodal profile reflects recurring camera and lens configurations rather than one optical system.
A downstream visual policy must recover hand-object geometry and motion across different projections, crops, distortions, and imaging pipelines.
This variation provides a form of natural sensor-domain randomization that may improve robustness when transferring to unseen cameras, deployment rigs, and robot viewpoints.

\textbf{Takeaway.} Open-AoE is differentiated not only by scale, but by diversity at three coupled levels: what people manipulate, where and how they act, and the consumer-camera domain through which every interaction is observed.
Together, these properties create a stronger basis for visually robust, cross-domain manipulation learning.

\textit{Scope and claim boundary.} Every distribution reported in this subsection is computed from the same random 100-hour sample. The downstream benefit of camera-domain diversity remains a training hypothesis that should be validated through controlled ablations.

\section{Open-AoE Toolchain}

The Open-AoE Toolchain turns synchronized egocentric video, hand geometry, camera motion, and action semantics into research assets that can be inspected, reconstructed, retargeted, and used for learning.
The released implementation, documentation, and model-integration recipes are available in the \href{https://github.com/ant-research/Open-AoE}{\raisebox{-0.15ex}{\includegraphics[height=1.05em]{logo/github-logo.pdf}}\hspace{0.25em}Open-AoE GitHub repository}.
Each AoE sample aligns undistorted RGB, camera intrinsics and trajectories, bilateral MANO reconstruction~\cite{mano2017}, validity masks, and atomic-action annotations on one timeline.
These signals form a common evidence layer, but no single action vector is appropriate for human-motion modeling, robot control, and visual-dynamics learning at the same time.
The toolchain therefore preserves the physical meaning of the source signals and selects a representation level according to the downstream problem.

\subsection{AoE-Visualization}

AoE-Visualization is the entry point for synchronized inspection and error diagnosis.
It uses the undistorted egocentric video as the canvas, overlays MANO meshes, 21-keypoint skeletons, future wrist trajectories, and atomic actions, and presents a world-frame 3D view and timeline alongside the image.
This joint view exposes reprojection error, missing-hand intervals, SLAM drift, and semantic-boundary misalignment before a downstream model can absorb them as apparent motion regularities.
The same visual grammar is used for dataset examples and for reviewing intermediate toolchain outputs.

\subsection{AoE-Reconstruct-Retarget}

AoE-Reconstruct-Retarget connects an egocentric observation to three complementary outputs: 4D hand-object assets, robot-executable trajectories, and robotized video, as summarized in Figure~\ref{fig:reconstruct-retarget}.
The outputs are coupled: reconstruction supplies object and contact references, retargeting supplies executable motion, and robotized video places that motion back into the real visual context.

\begin{center}
    \centering
    \includegraphics[width=\textwidth]{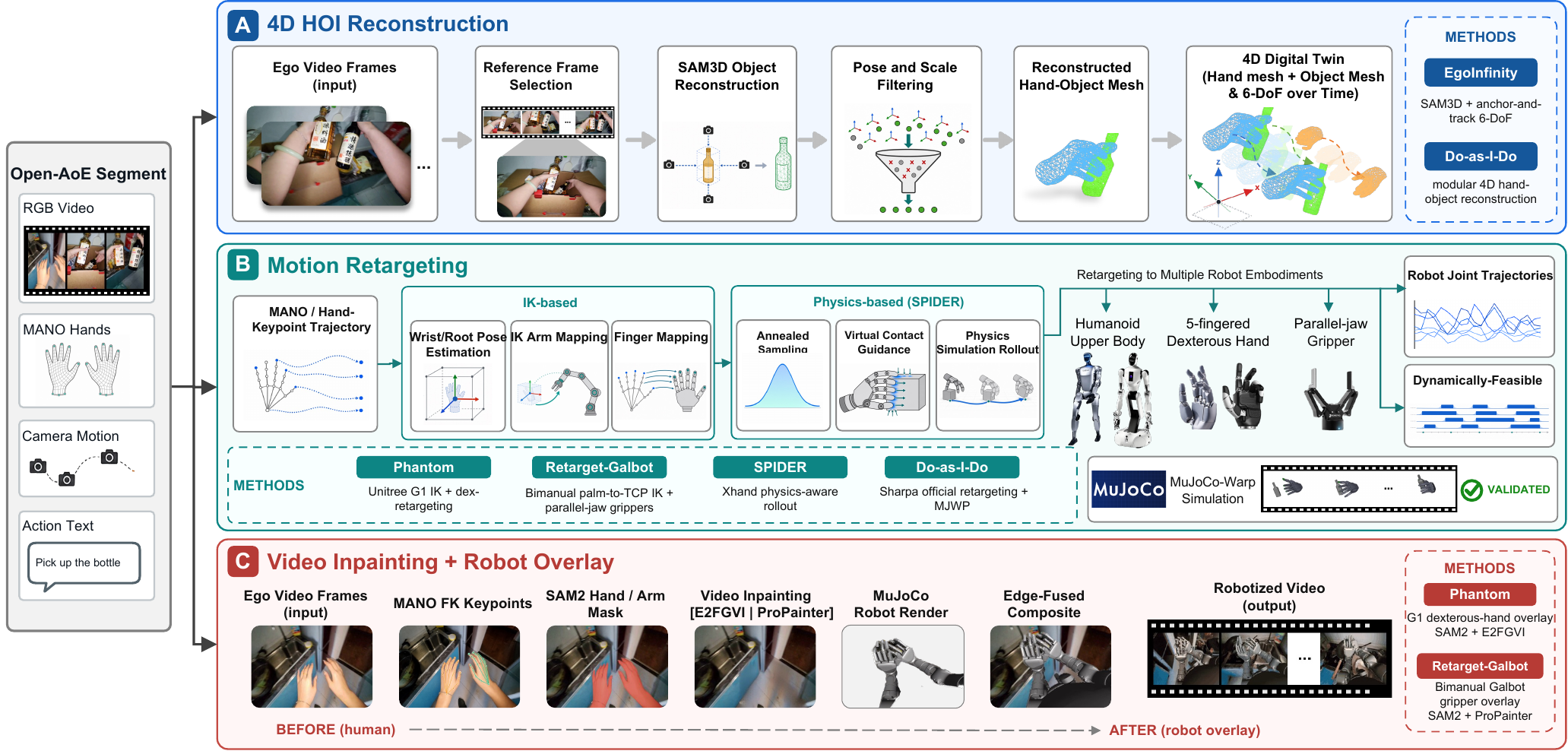}
    \captionof{figure}{\textbf{AoE-Reconstruct-Retarget.} A smartphone-captured egocentric video is projected into 4D hand-object assets, robot-usable motion, and robotized video. The repository integrates reconstruction interfaces for EgoInfinity and Do-as-I-Do, multiple retargeting backends for G1, Galbot/Galaxea, Sharpa, and XHand, and simulation and egoview synthesis for result validation.}
    \label{fig:reconstruct-retarget}
\end{center}

\paragraph{4D hand-object reconstruction.} EgoInfinity~\cite{egoinfinity2026} and Do-as-I-Do~\cite{doasido2026} lift monocular video into time-varying hand geometry, object pose, and interaction assets.
Here, AoE contributes more than additional RGB frames: calibrated intrinsics and metric-scale camera trajectories help separate observer motion from scene motion, MANO supplies hand-scale and pose priors, and action segments delimit the temporal support of an interaction.
When combined with object 6-DoF estimates, the timing of finger approach, grasp, and release can provide contact candidates, object-tracking constraints, and priors over manipulation phases.
This allows large-scale egocentric data to convey knowledge about object affordances and hand-object coordination rather than only scene appearance.

\paragraph{Cross-embodiment retargeting.} This route asks how motion demonstrated by a human can become motion executable by a target robot.
Phantom combines arm inverse kinematics with dex-retargeting~\cite{dexretarget} to map camera-local wrist and finger motion to Unitree G1 with Dex3 or Inspire hands.
Retarget Galbot converts bimanual palm targets into Galbot/Galaxea TCP trajectories and gripper control, while Retarget Lab organizes comparable kinematic and physics-aware routes including EgoInfinity/G1, Do-as-I-Do/Sharpa, and SPIDER/XHand~\cite{spider2025}.
AoE's bilateral MANO trajectories, validity masks, and atomic-action boundaries jointly provide supervision for wrist paths, finger synergies, bimanual role assignment, and grasp timing.
Compared with isolated end-effector poses, these signals are better suited to learning cross-embodiment motion priors because they preserve the structure among reach, contact, manipulation, and release.

\paragraph{Robotized video.} This route removes the human arm and composites a simulated robot back into the original egocentric scene.
Phantom and Retarget Galbot use MANO keypoints as geometric prompts for segmentation and combine video inpainting with MuJoCo rendering to produce temporally aligned robot overlays.
Because it preserves the real environment, manipulated object, illumination, and task progression while changing the actor appearance, it can provide paired signals for human-to-robot visual-domain transfer, retargeting diagnosis, and robot-scene consistency learning.
Its value is to create aligned observations of the same interaction under different embodiments, not to treat synthesized video as a substitute for real robot execution.

\subsection{AoE-Training-Ready}

AoE-Training-Ready adapts the same synchronized source signals to model-specific action semantics rather than claiming a universal training format; Figure~\ref{fig:training-ready} summarizes this representation spectrum and its downstream model connections.
Multiple representations are necessary because human-motion learning, robot control, and video generation preserve different invariants, as summarized in Table~\ref{tab:aoe_action_representations}: human modeling benefits from finger articulation, cross-embodiment control needs end-effector or joint targets, and egocentric dynamics must account for camera self-motion.

\begin{center}
    \centering
    \includegraphics[width=\textwidth]{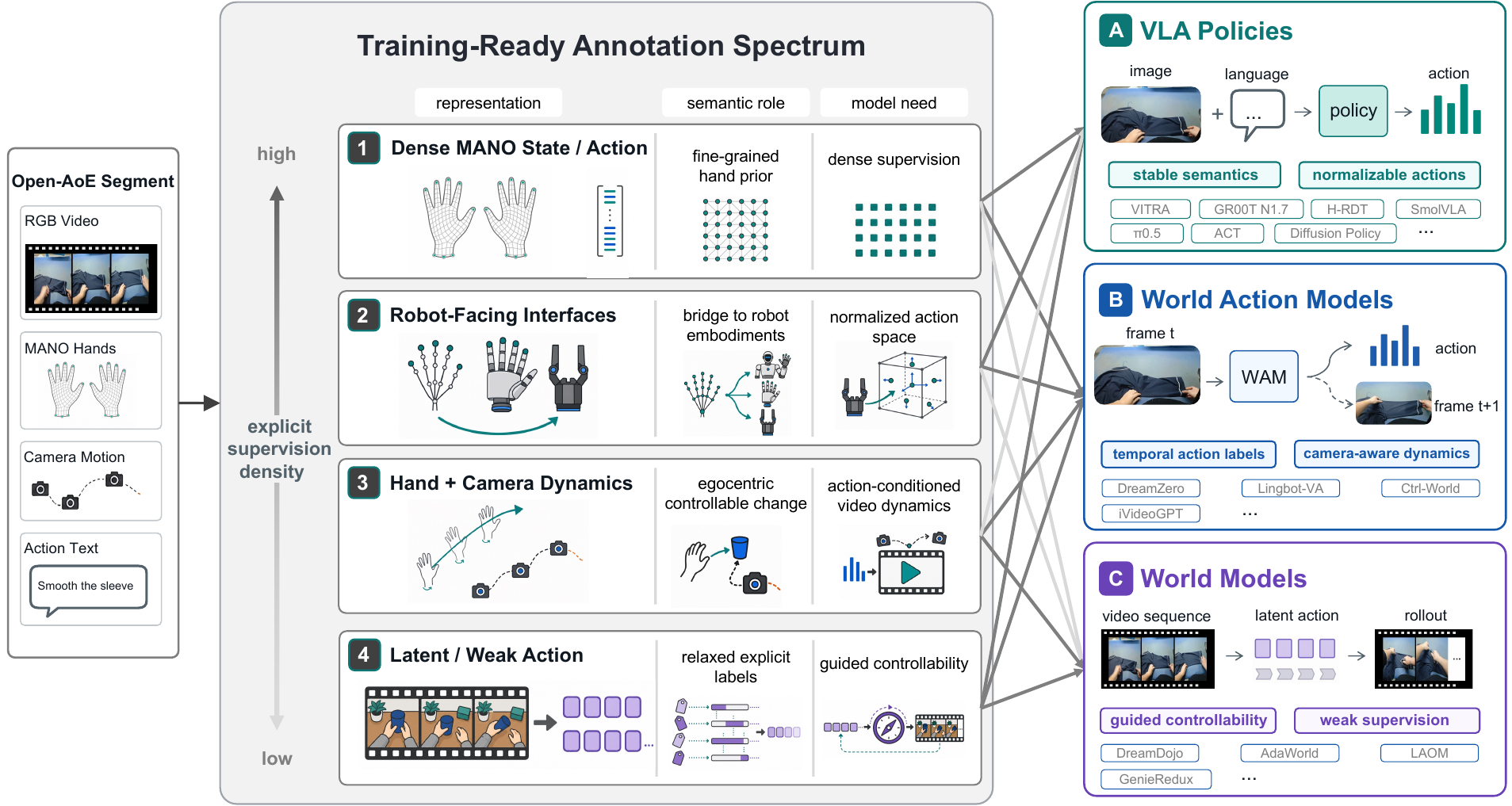}
    \captionof{figure}{\textbf{AoE-Training-Ready.} A synchronized AoE segment is projected into action representations with different supervision densities and semantic roles. Integrated recipes connect dense MANO, robot-facing, hand-plus-camera, and latent or weak action representations to Vision-Language-Action policies, World Action Models, and World Models.}
    \label{fig:training-ready}
\end{center}

For recipes that require dense hand articulation, Open-AoE concatenates two 55D per-hand vectors into a 110D representation: $2\!\times\![\mathrm{valid}(1)+\mathrm{wrist\ translation}(3)+\mathrm{wrist\ axis\mbox{-}angle}(3)+\mathrm{velocity}(3)+\mathrm{MANO\ pose}(45)]$.
Translations and finite-difference velocities are expressed in meters and meters per second in the current-frame OpenCV camera coordinates ($x$ right, $y$ down, $z$ forward), while rotations are represented in radians; because this frame moves with the wearer, the velocity contains both hand motion and camera ego-motion rather than purely inertial wrist motion.

\noindent\begin{minipage}{\textwidth}
\centering
\captionof{table}{Action representations exposed by AoE-Training-Ready. Dimensions are per-frame vectors defined by Open-AoE converters; citations identify integrated downstream methods rather than the origin of the layouts. Recipes may normalize vectors or concatenate a future horizon. The two 20D interfaces are not interchangeable.}
\label{tab:aoe_action_representations}
\footnotesize
\setlength{\tabcolsep}{3.5pt}
\begin{tabularx}{\textwidth}{>{\raggedright\arraybackslash}p{2.0cm} >{\raggedright\arraybackslash}p{4.45cm} >{\raggedright\arraybackslash}p{3.25cm} >{\raggedright\arraybackslash}X}
\toprule
Interface & Layout & Frame and units & Learning role / representative consumers \\
\midrule
110D dense MANO~\cite{mano2017} & $2\!\times\![\mathrm{valid}(1)+\mathrm{wrist}\ xyz(3)+\mathrm{wrist}\ aa(3)+\mathrm{velocity}(3)+\mathrm{MANO}(45)]$ & Current-camera frame; m, rad, m/s, binary validity & Dense hand state and next-state targets; Open-AoE adapters for LeRobot~\cite{cadene2026lerobot} and world-action recipes. \\
\midrule
48D wrist--fingertip & $2\!\times\![\mathrm{wrist}\ xyz(3)+\mathrm{rot6D}(6)+5\ \mathrm{fingertips}\ xyz(15)]$ & SLAM world frame; positions in m, rot6D unitless & Geometry-preserving target used by the Open-AoE H-RDT adapter~\cite{hrdt2025}. \\
\midrule
62D Sharpa & $L/R\ \mathrm{wrist\ EEF}(9)+L/R\ \mathrm{Sharpa\ joints}(22)$ & World-frame wrist in m + rot6D; robot joints in rad & Open-AoE GR00T N1.7 adapter~\cite{grootn172026}; actions encode wrist EEF relatively and joints absolutely. \\
\midrule
20D GR00T gripper & $L/R\ \mathrm{wrist\ EEF}(9)+L/R\ \mathrm{gripper}(1)$ & World-frame wrist in m + rot6D; gripper in $[0,1]$ & Lower-DoF Open-AoE GR00T N1.7 adapter~\cite{grootn172026}; wrist EEF is relative and gripper absolute. \\
\midrule
22D state / 20D--26D ego action & State: $2\!\times\![xyz(3)+\mathrm{rot6D}(6)+\mathrm{grip}(1)]+2\ \mathrm{valid}$; action drops validity and optionally appends camera $\Delta t(3)+\Delta r_{aa}(3)$ & Hand in current-camera frame; m and unitless rot6D/grip; camera delta in m and rad & Open-AoE adapters for SmolVLA~\cite{smolvla2025}, iVideoGPT~\cite{ivideogpt2024}, LAOM~\cite{laom2025}, and guided GenieRedux~\cite{genieredux2024}. \\
\bottomrule
\end{tabularx}
\end{minipage}

\paragraph{Policy learning and cross-embodiment control.} This route compresses egocentric human motion into targets that a policy can predict and a robot can execute.
VITRA~\cite{vitra2025}, H-RDT~\cite{hrdt2025}, GR00T N1.7~\cite{grootn172026}, SmolVLA~\cite{smolvla2025}, and LeRobot-based~\cite{cadene2026lerobot} $\pi_{0.5}$, ACT, and Diffusion Policy recipes consume dense MANO, wrist--fingertip, robot-joint, or hand-as-EEF targets according to their control abstraction.
Atomic-action language supplies task intent, framewise hand geometry supplies motion realization, bilateral trajectories supply coordination structure, and long-tailed objects and behaviors in natural scenes supply affordance and behavioral diversity.
Together, these signals support a hierarchy from observing and describing an interaction to predicting embodiment-constrained motion, without requiring robot joint labels for every human video.

\paragraph{World-action and action-conditioned video learning.} This route models how the next action or future visual state changes given the current observation and an action condition.
DreamZero~\cite{dreamzero2026}, LingBot-VA~\cite{lingbova2026}, Ctrl-World~\cite{ctrlworld2025}, and iVideoGPT~\cite{ivideogpt2024} use either dense next-hand states or compact hand-plus-camera actions to learn action prediction and visual dynamics.
Large image motion in egocentric video may arise from hand-object interaction or from motion of the head or phone; because AoE synchronizes hand and camera trajectories, these two causal sources can be represented as separate conditions or controlled variables.
This supplies direct signals for controllable visual prediction, for testing whether a model genuinely uses its action condition, and for learning action representations that are robust to a moving camera.

\paragraph{Latent-action and world-model learning.} This route discovers action variables from visual change and uses limited structured supervision to determine what those variables should represent.
GenieRedux~\cite{genieredux2024} and AdaWorld~\cite{adaworld2025} learn latent actions from videos or frame pairs, LAOM~\cite{laom2025} uses limited hand-action supervision to resist camera ego-motion as a strong distractor, and DreamDojo~\cite{dreamdojo2026} injects explicit MANO motion into its action condition.
AoE therefore contributes three forms of knowledge: scene dynamics from unlabeled video, nuisance-factor identification from camera trajectories, and controllable weak supervision from MANO motion and action segments.
The same data can move from discovering latent factors of change, to grounding those factors as hand actions, to controlling future video with explicit motion, thereby connecting unsupervised world modeling with embodied control.

Together, AoE-Visualization, AoE-Reconstruct-Retarget, and AoE-Training-Ready connect data inspection, interaction reconstruction, cross-embodiment conversion, and model learning in one traceable path.
The central design is not the number of interfaces, but the reuse of the same egocentric evidence at geometric, action, and visual-dynamics levels.

\section{Dataset Analysis}
\label{sec:dataset-analysis}

To assess whether Open-AoE exhibits a distinctive data profile beyond dataset scale, we compare it with OpenEgo~\cite{openegocentric2025}, EgoDex~\cite{egodex2025}, and EgoXtreme~\cite{egoxtreme2026} across four complementary dimensions: visual representation diversity, semantic and temporal supervision, annotation consistency, and downstream training utility. %
Approximately 100 hours of data are included for Open-AoE, OpenEgo, and EgoDex, whereas EgoXtreme is evaluated using its entire available release because its total duration is substantially smaller. %
The analysis therefore focuses not only on the breadth of the visual and semantic distributions, but also on the completeness, consistency, and practical usability of the associated supervision.

\begin{figure*}[t]
    \centering
    \includegraphics[width=\linewidth]{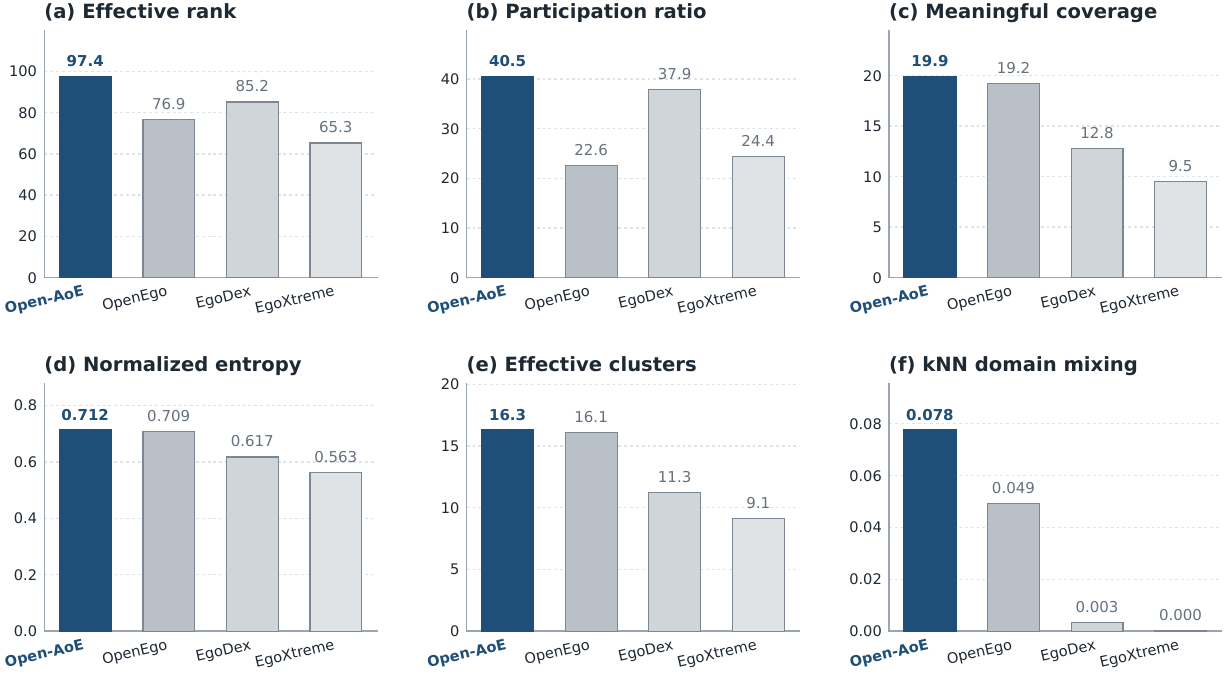}
    \caption{High-dimensional CLIP feature analysis. Bars report the mean over 3,000 balanced random subsamples, with 2,000 embeddings drawn from each dataset in every trial. Open-AoE has the highest mean in all six metrics.}
    \label{fig:clip_six_metrics}
\end{figure*}

\subsection{High-Dimensional Visual Diversity}

We use CLIP~\cite{radford2021clip} image embeddings to compare the visual feature distributions of the four datasets. %
Visual frames are sampled from each video sequence and encoded using the CLIP image encoder. %
Each embedding is then $\ell_2$-normalized, yielding $\mathbf{z}_i\in\mathbb{R}^{D}$, where $i$ indexes a sampled visual observation and $D$ is the CLIP feature dimension. %
To control for unequal dataset sizes, we conduct 3,000 balanced random trials, each sampling 2,000 embeddings without replacement from every dataset. %
All reported values are averaged across these trials.

We first examine the covariance spectrum of each dataset in the original normalized CLIP space. %
For dataset $s$, let $\lambda_{s,j}$ denote the $j$-th eigenvalue of its embedding covariance matrix, where $j=1,\ldots,D$, and let $p_{s,j}$ be the corresponding proportion of total variance. %
Effective Rank is the exponential entropy of the normalized covariance spectrum, whereas Participation Ratio measures its inverse concentration:
\begin{equation}
p_{s,j}=\frac{\lambda_{s,j}}{\sum_r\lambda_{s,r}},
\qquad
R_{\mathrm{eff}}^{(s)}
=\exp\!\left(-\sum_j p_{s,j}\log p_{s,j}\right),
\qquad
R_{\mathrm{PR}}^{(s)}
=\frac{\left(\sum_j\lambda_{s,j}\right)^2}
{\sum_j\lambda_{s,j}^{2}}.
\end{equation}
Both metrics estimate the effective number of feature directions that carry substantial variance. %
Higher values therefore indicate that visual variation is distributed across a larger set of active directions rather than being concentrated in a few dominant modes.

We next measure how broadly and evenly each dataset occupies a shared visual feature partition. %
In each trial, the pooled balanced embeddings are transformed using a common PCA representation and partitioned into a shared codebook of $K=50$ cells. %
Let $n_{s,c}$ denote the number of embeddings from dataset $s$ assigned to cell $c$, and let $q_{s,c}$ denote the corresponding occupancy proportion. %
Meaningful Coverage counts cells containing at least $\tau$ samples, while Normalized Cluster Entropy and Effective Clusters quantify the balance of the occupancy distribution:
\begin{equation}
\begin{aligned}
q_{s,c}
&=\frac{n_{s,c}}{\sum_{u=1}^{K}n_{s,u}},
&
C_{s,\tau}
&=\sum_{c=1}^{K}\mathbf{1}\!\left[n_{s,c}\ge\tau\right],
\\
H_{\mathrm{norm}}^{(s)}
&=-\frac{\sum_{c=1}^{K}q_{s,c}\log q_{s,c}}{\log K},
&
N_{\mathrm{eff}}^{(s)}
&=\exp\!\left(-\sum_{c=1}^{K}q_{s,c}\log q_{s,c}\right).
\end{aligned}
\end{equation}
Here, $C_{s,\tau}$ denotes the number of visual regions with sufficiently stable sample support, with $\tau=5$ used throughout our experiments. %
$H_{\mathrm{norm}}^{(s)}$ measures how evenly the samples are distributed across the shared codebook, while $N_{\mathrm{eff}}^{(s)}$ expresses the same entropy as an equivalent number of uniformly occupied cells. %
Higher values indicate broader and more balanced coverage of the shared visual feature space.

Finally, we compute kNN Domain Mixing directly in the original normalized CLIP space, without codebook discretization. %
Let $y_i$ denote the dataset identity of embedding $i$, let $N_s$ be the number of sampled embeddings from dataset $s$, and let $\mathcal{N}_k(i)$ denote the set of its $k$ nearest neighbors in the pooled sample. %
The metric is defined as
\begin{equation}
M_{\mathrm{kNN}}^{(s)}
=
1-\frac{1}{N_s}
\sum_{i:y_i=s}
\frac{1}{k}
\sum_{j\in\mathcal{N}_k(i)}
\mathbf{1}\!\left[y_j=s\right],
\end{equation}
where $\mathbf{1}[\cdot]$ is the indicator function and $k=20$. %
The inner term measures the proportion of same-dataset neighbors around an embedding; therefore, $M_{\mathrm{kNN}}^{(s)}$ is the average proportion of neighboring samples originating from other datasets. %
A higher value indicates stronger local connectivity to a broader range of visual domains.

As shown in Fig.~\ref{fig:clip_six_metrics}, across the 3,000 balanced trials, Open-AoE achieves the highest mean in all six metrics: Effective Rank ($97.43$), Participation Ratio ($40.47$), Meaningful Coverage ($19.89$), Normalized Cluster Entropy ($0.712$), Effective Clusters ($16.29$), and kNN Domain Mixing ($0.0775$). %
The largest advantages are observed in Effective Rank and kNN Domain Mixing, indicating that Open-AoE spans a larger number of active feature directions while maintaining stronger local connectivity to other visual domains. %
Moreover, Open-AoE ranks first in Effective Rank, Participation Ratio, and kNN Domain Mixing in all 3,000 trials. %
Its advantages in the coverage- and entropy-based metrics are smaller, particularly relative to OpenEgo, but their repeated-sampling means remain the highest among all four datasets.

\begin{figure*}[t]
    \centering
    \includegraphics[width=\linewidth]{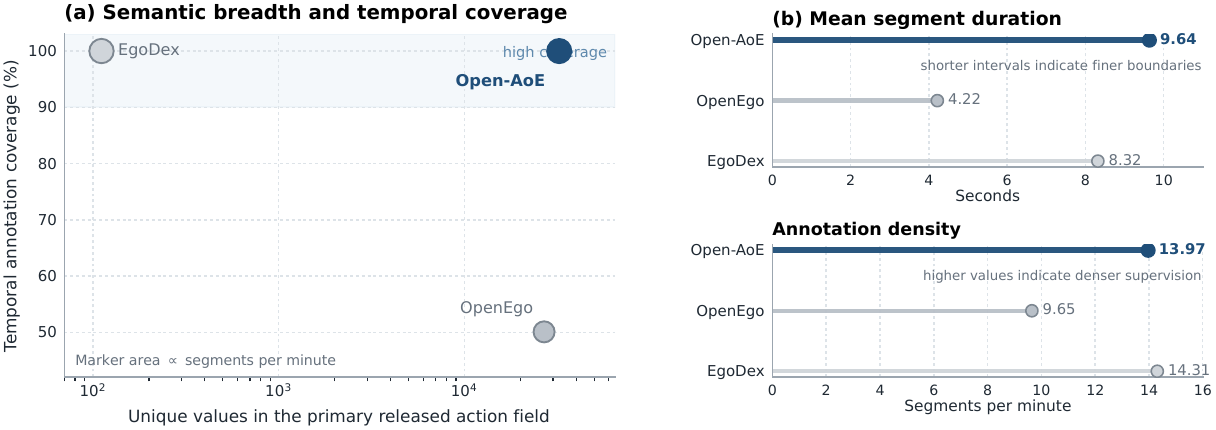}
    \caption{
    \textbf{Semantic breadth and temporal annotation characteristics across datasets.}
    (a) Number of unique entries in the primary released action-annotation field versus temporal annotation coverage, with marker area proportional to annotation density, measured as the number of annotated segments per minute.
    (b) Mean segment duration and annotation density for each dataset.
    Because the datasets use different native annotation schemas, vocabulary size reflects the breadth of the released semantic supervision rather than a controlled vocabulary comparison under a shared ontology.
    }
    \label{fig:temporal_semantic}
\end{figure*}

\subsection{Semantic Breadth and Temporal Completeness}

We next examine the semantic breadth and temporal structure of the released action annotations. %
Temporal coverage is defined as the fraction of the evaluated sequence timeline covered by at least one action segment, measuring the continuity of semantic supervision over the underlying video. %
The released action vocabulary is quantified by the number of unique entries in each dataset's primary action-annotation field. %
Because the datasets follow different native annotation schemas, this measure reflects the breadth of semantic supervision exposed by each release rather than a controlled comparison under a shared action ontology. %
We further characterize annotation structure using mean segment duration and annotation density, defined as the number of annotated segments per minute. %
The former captures the typical temporal extent of individual action annotations, whereas the latter reflects their frequency along the video timeline. %
Taken together, these metrics capture complementary aspects of supervision: vocabulary size reflects semantic breadth, temporal coverage measures annotation completeness, and segment duration and density characterize the granularity of localized action annotation.

As illustrated in Fig.~\ref{fig:temporal_semantic}, Open-AoE combines broad natural-language supervision with near-complete temporal coverage. %
Its primary action field comprises 32,407 distinct natural-language descriptions, compared with 26,864 native labels in OpenEgo and 111 categorical action classes in EgoDex. %
Beyond these descriptions, Open-AoE includes structured semantic fields with 8,030 object strings, 175 action verbs, and 135 scene labels, enabling each action interval to be represented at multiple levels of semantic abstraction. %
Its annotations cover 99.99\% of the evaluated timeline, substantially exceeding the 50.1\% coverage of OpenEgo, while retaining a high annotation density of 13.97 segments per minute. %
EgoDex achieves complete clip-level label coverage and a slightly higher segment density, but relies on a substantially smaller categorical action vocabulary. %
Overall, Open-AoE combines open-vocabulary semantic breadth with near-continuous temporal alignment and densely localized action supervision, without compromising annotation completeness.

\begin{figure*}[t]
    \centering
    \includegraphics[width=1.0\linewidth]{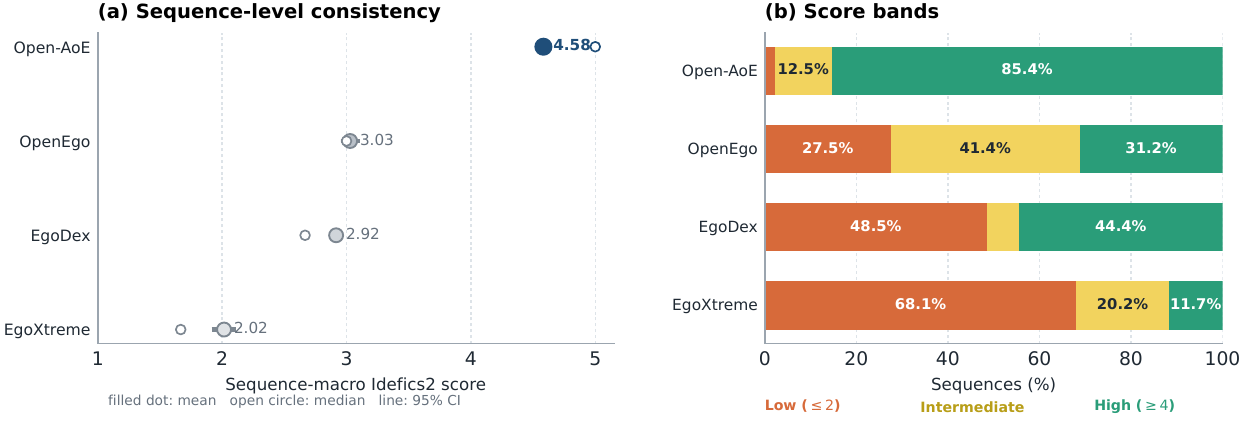}
    \caption{
    \textbf{Sequence-level image-annotation consistency across datasets.}
    (a) Sequence-macro means, sequence-level medians, and 95\% bootstrap confidence intervals for the mean Idefics2 consistency score.
    (b) Fractions of sequences in the low ($\leq 2$), intermediate ($2<\cdot<4$), and high ($\geq 4$) score ranges.
    }
    \label{fig:vlm_consistency}
\end{figure*}

\subsection{Image-Annotation Consistency}

We assess whether the released action annotations are supported by the corresponding visual content using Idefics2~\cite{laurencon2024idefics2} as a shared vision-language evaluator. %
Let $a_{q,t}\in[1,5]$ denote the consistency score assigned to the $t$-th sampled frame of sequence $q$. %
The sequence-macro score is defined as \begin{equation} S_{\mathrm{macro}} = \frac{1}{Q} \sum_{q=1}^{Q} \left( \frac{1}{T_q} \sum_{t=1}^{T_q} a_{q,t} \right), \end{equation}
where $Q$ is the number of evaluated sequences and $T_q$ is the number of sampled frames in sequence $q$. %
Frame-level scores are first averaged within each sequence and then across sequences, thereby assigning equal weight to each sequence and preventing longer recordings from dominating the dataset-level estimate. %
Higher scores indicate stronger agreement between the released annotations and the sampled visual evidence.

As shown in Fig.~\ref{fig:vlm_consistency}, Open-AoE achieves a sequence-macro score of $4.583/5$, substantially exceeding those of OpenEgo ($3.029$), EgoDex ($2.916$), and EgoXtreme ($2.015$). %
Its 95\% bootstrap confidence interval is $[4.557,\,4.608]$. %
This advantage is also evident in the score distribution: 85.4\% of Open-AoE sequences attain a mean consistency score of at least 4, whereas only 2.2\% score at or below 2. %
The higher aggregate score therefore reflects a broad concentration of sequences near the upper end of the scoring range rather than a small subset of exceptionally high-scoring cases. %
Because the datasets differ in their native annotation schemas and levels of textual detail, these results should be interpreted as a cross-dataset audit of image-annotation support under a shared evaluator rather than as a fully controlled benchmark of annotation accuracy.

\begin{figure*}[t]
    \centering
    \includegraphics[width=\linewidth]{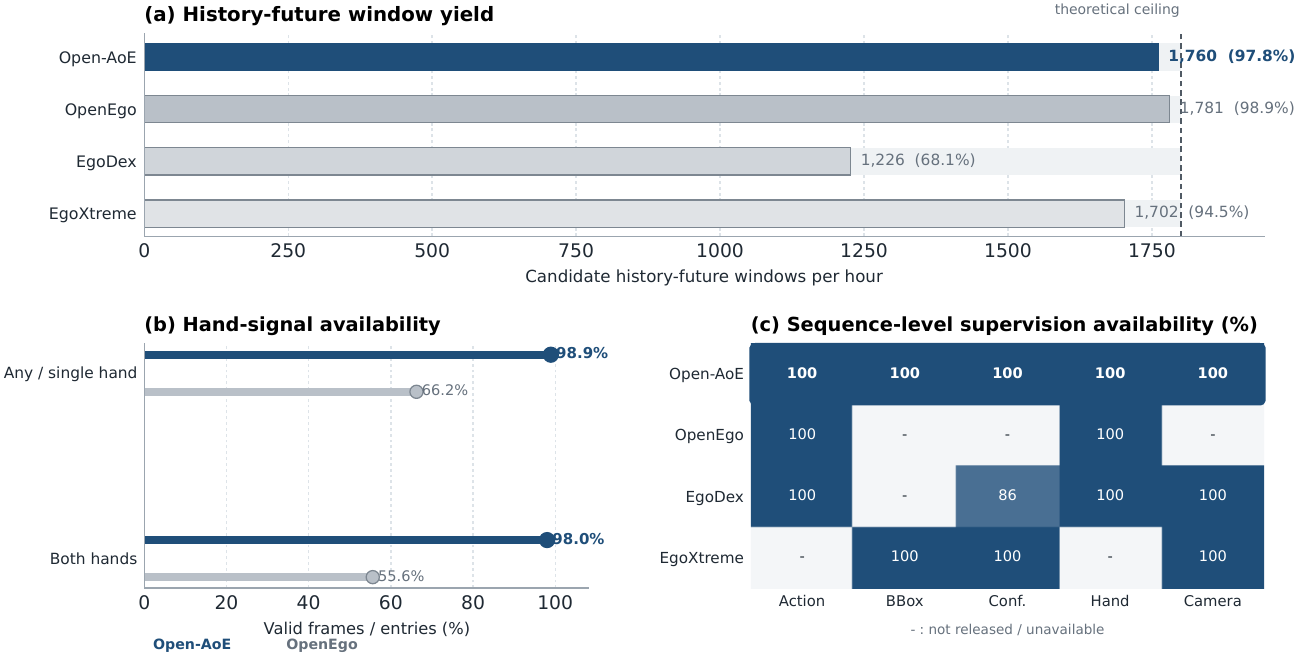}
    \caption{
    \textbf{Training-sample yield and multimodal supervision availability across datasets.}
    (a) Candidate history-future windows obtained per hour using 2~s of history, 2~s of future context, and a 2~s stride, together with the retention rate relative to the no-boundary ceiling of 1,800 windows per hour.
    (b) Availability of native hand signals, distinguishing entries with at least one valid hand from those with valid signals for both hands.
    (c) Sequence-level availability of action annotations, bounding boxes, annotation confidence, hand pose, and camera pose; ``-'' indicates that the corresponding field is not released or unavailable.
    }
    \label{fig:training_utility}
\end{figure*}

\subsection{Training-Window Retention and Multimodal Completeness}

We finally assess how efficiently the released sequences can be converted into temporally aligned training samples and whether the principal supervision signals required by downstream embodied-learning pipelines are jointly available. %
For a sequence $q$ of duration $L_q$, we construct candidate samples using a history horizon of $h=2$~s, a future horizon of $f=2$~s, and a temporal stride of $\Delta=2$~s. %
The number of candidate windows is defined as
\begin{equation}
N_q
=
\max\!\left(
0,\,
\left\lfloor \frac{L_q-h-f}{\Delta} \right\rfloor + 1
\right).
\end{equation}

The history-future window yield and normalized retention are then defined as
\begin{equation}
Y_{\mathrm{window}}
=
\frac{\sum_q N_q}{\sum_q L_q/3600},
\qquad
\eta_{\mathrm{window}}
=
\frac{Y_{\mathrm{window}}}{3600/\Delta},
\end{equation}
where $\eta_{\mathrm{window}}$ measures retention relative to the no-boundary ceiling of $3600/\Delta=1{,}800$ windows per hour. %
This metric isolates losses induced by sequence boundaries and temporal discontinuities, excluding task-specific filtering imposed by individual downstream models. %
We additionally evaluate native hand-signal availability using the released validity information, distinguishing between entries with at least one valid hand and those with valid signals for both hands. %
At the sequence level, we record the availability of five principal supervision modalities: action annotations, bounding boxes, annotation confidence, hand pose, and camera pose.

As shown in Fig.~\ref{fig:training_utility}, Open-AoE yields approximately 1,760 candidate history-future windows per hour, corresponding to 97.8\% of the theoretical ceiling. %
This small gap from the no-boundary limit indicates that its temporal continuity enables dense extraction of fixed-context training samples with minimal loss at sequence boundaries. %
Open-AoE also exhibits substantially higher native hand-signal availability than OpenEgo, with 98.93\% of evaluated entries containing at least one valid hand signal and 98.02\% containing valid signals for both hands. %
Moreover, it is the only evaluated release in which all five supervision modalities are available with near-universal sequence-level coverage. %
Bounding-box quality is similarly high, with 100\% valid boxes, 98.39\% fully contained within the image boundaries, a mean confidence of 0.947, and a 10th-percentile confidence of 0.950. %
These results show that Open-AoE combines high temporal sample retention with dense, jointly available semantic, geometric, hand-motion, and camera-motion supervision, thereby reducing the modality filtering and temporal alignment required before downstream VLA, world-model, and human-to-robot training.

\section{Conclusion}

We presented Open-AoE, an open data infrastructure that couples approximately 2,000 hours of real-world egocentric manipulation data captured with consumer smartphones with a complete path from data to models.
On the production side, the capture and processing pipeline turns raw smartphone video into structured samples aligned across vision, language, hand motion, camera trajectories, and action boundaries; on the consumption side, the open-source toolchain supports synchronized visualization, 4D reconstruction, cross-embodiment retargeting, and training representations for Vision-Language-Action policies, World Action Models, and World Models.
Its broad coverage of scenes, actions, interaction objects, participants, device models, and fields of view allows this infrastructure to connect low-cost real-world collection with multiple embodied-learning paradigms.

Open-AoE is intended not as a static data package, but as an open ecosystem co-developed by data contributors, universities, research institutions, robot developers, and model teams.
We invite the community to contribute new real-world scenes and tasks, reconstruction and retargeting methods, training adapters, and evaluation results, allowing data, tools, and models to evolve together in an open loop.
Our goal is to make Open-AoE the lowest-barrier egocentric data infrastructure for embodied intelligence---making egocentric data easy for anyone to collect and use, so that data is no longer the bottleneck for embodied foundation models.

\clearpage
\section*{Contributors}

\noindent The following contributors are listed by module contribution.

\subsection*{Dataset and Visualization}
Zishuo Li$^{*,1}$, Qingze Guan$^{*,5}$, Zhengxing Wu$^{*,7}$, Wanke Zhan$^{*,1}$, Yang Sun$^{*,1}$, Zhiyi Huang$^{*,1}$, Zitong Shan$^{*,1}$

\subsection*{Toolchain and Experiments}
Bowen Yang$^{*,1}$, Changtao Miao$^{*,1}$, Zhenchao Jin$^{*,3}$, Jiadong Hong$^{*,2}$, Taowen Wang$^{*,4}$, Yushi Feng$^{*,3}$, You Liu$^{*,6}$, Yibo Wang$^{*,6}$, Yifan Yang$^{*,7}$, Zhaowen Zhou$^{*,1}$, Man Luo$^{*,1}$, Hao Cheng$^{*,8}$

\subsection*{Project Leaders}
Zishuo Li$^{1}$, Bowen Yang$^{1}$, Changtao Miao$^{\ddag,1}$, Kai Zhu$^{\ddag,1}$, Hao Chen$^{\ddag,2}$

\subsection*{Supervisors}
Bo Zhang$^{1}$, Jianshu Li$^{1}$, Jiansheng Cai$^{1}$, Guocai Yao$^{6}$, Jize Zhang$^{8}$, Chenhao Lin$^{9}$, Renjing Xu$^{4}$, Lequan Yu$^{3}$, Chao Shen$^{9}$, Chunhua Shen$^{2}$, Zhe Li$^{1}$

\vspace{3mm}
\noindent $^{*}$Equal contribution. $^{\ddag}$Corresponding authors.

\section*{Organizations}
\noindent $^{1}$Ant Digital Technology, Ant Group

\noindent $^{2}$Zhejiang University

\noindent $^{3}$The University of Hong Kong

\noindent $^{4}$The Hong Kong University of Science and Technology (Guangzhou)

\noindent $^{5}$National University of Singapore

\noindent $^{6}$Beijing Academy of Artificial Intelligence

\noindent $^{7}$University of Chinese Academy of Sciences

\noindent $^{8}$The Hong Kong University of Science and Technology

\noindent $^{9}$Xi'an Jiaotong University

\clearpage
\bibliographystyle{unsrtnat}
\bibliography{main}

\begin{thebibliography}{45}
\providecommand{\natexlab}[1]{#1}
\providecommand{\url}[1]{\texttt{#1}}
\expandafter\ifx\csname urlstyle\endcsname\relax
  \providecommand{\doi}[1]{doi: #1}\else
  \providecommand{\doi}{doi: \begingroup \urlstyle{rm}\Url}\fi

\bibitem[Grauman et~al.(2022)]{grauman2022ego4d}
Kristen Grauman et~al.
\newblock {Ego4D}: Around the world in 3,000 hours of egocentric video.
\newblock In \emph{Proceedings of the IEEE/CVF Conference on Computer Vision
  and Pattern Recognition}, 2022.
\newblock URL
  \url{https://openaccess.thecvf.com/content/CVPR2022/html/Grauman_Ego4D_Around_the_World_in_3000_Hours_of_Egocentric_Video_CVPR_2022_paper.html}.

\bibitem[Damen et~al.(2022)Damen, Doughty, Farinella, Furnari, Kazakos, Ma,
  Moltisanti, Munro, Perrett, Price, and Wray]{damen2022rescaling}
Dima Damen, Hazel Doughty, Giovanni~Maria Farinella, Antonino Furnari,
  Evangelos Kazakos, Jian Ma, Davide Moltisanti, Jonathan Munro, Toby Perrett,
  Will Price, and Michael Wray.
\newblock Rescaling egocentric vision: Collection, pipeline and challenges for
  {EPIC-KITCHENS-100}.
\newblock \emph{International Journal of Computer Vision}, 130\penalty0
  (1):\penalty0 33--55, 2022.

\bibitem[Hoque et~al.(2025)Hoque, Huang, Yoon, Sivapurapu, and
  Zhang]{egodex2025}
Ryan Hoque, Peide Huang, David~J. Yoon, Mouli Sivapurapu, and Jian Zhang.
\newblock {EgoDex}: Learning dexterous manipulation from large-scale egocentric
  video.
\newblock \emph{arXiv preprint arXiv:2505.11709}, 2025.
\newblock URL \url{https://arxiv.org/abs/2505.11709}.

\bibitem[Li et~al.(2026{\natexlab{a}})Li, Wei, Luo, Xiao, Bai, Zhou, Zou, Gui,
  Wen, Zhang, Chen, Pan, Liu, Wang, An, Li, Jin, Zhang, Wang, Wei, Huang, Liu,
  Li, Zhang, Li, Gong, Cao, Liang, and Lin]{egolive2026}
Yihang Li, Xuelong Wei, Jingzhou Luo, Yingjing Xiao, Yibo Bai, Guangyuan Zhou,
  Teng Zou, Chenguang Gui, Jiajun Wen, He~Zhang, Kangliang Chen, Xing Pan,
  Shuaiyan Liu, Daming Wang, Tao An, Jiayi Li, Shibo Jin, Wanwan Zhang, Tianyu
  Wang, Boren Wei, Zhixuan Huang, Fangsheng Liu, Ruodai Li, Hui Zhang, Anson
  Li, Yicheng Gong, Peng Cao, Jiaming Liang, and Liang Lin.
\newblock {EgoLive}: A large-scale egocentric dataset from real-world human
  tasks.
\newblock \emph{arXiv preprint arXiv:2604.23570}, 2026{\natexlab{a}}.
\newblock URL \url{https://arxiv.org/abs/2604.23570}.

\bibitem[Jawaid and Xiang(2025)]{openegocentric2025}
Ahad Jawaid and Yu~Xiang.
\newblock {OpenEgo}: A large-scale multimodal egocentric dataset for dexterous
  manipulation.
\newblock \emph{arXiv preprint arXiv:2509.05513}, 2025.
\newblock URL \url{https://arxiv.org/abs/2509.05513}.

\bibitem[Zheng et~al.(2026)Zheng, Niu, Xie, Wang, Xu, Jiang, Casta{\~n}eda, Hu,
  Tan, Fu, Darrell, Huang, Zhu, Xu, and Fan]{egoscale2026}
Ruijie Zheng, Dantong Niu, Yuqi Xie, Jing Wang, Mengda Xu, Yunfan Jiang,
  Fernando Casta{\~n}eda, Fengyuan Hu, You~Liang Tan, Letian Fu, Trevor
  Darrell, Furong Huang, Yuke Zhu, Danfei Xu, and Linxi Fan.
\newblock {EgoScale}: Scaling dexterous manipulation with diverse egocentric
  human data.
\newblock \emph{arXiv preprint arXiv:2602.16710}, 2026.
\newblock URL \url{https://arxiv.org/abs/2602.16710}.

\bibitem[Lin et~al.(2024)Lin, Hu, Sheng, Wen, You, and Gao]{lindata}
Fanqi Lin, Yingdong Hu, Pingyue Sheng, Chuan Wen, Jiacheng You, and Yang Gao.
\newblock Data scaling laws in imitation learning for robotic manipulation.
\newblock \emph{arXiv preprint arXiv:2410.18647}, 2024.
\newblock URL \url{https://arxiv.org/abs/2410.18647}.

\bibitem[Khazatsky et~al.(2024)]{droid2024}
Alexander Khazatsky et~al.
\newblock {DROID}: A large-scale in-the-wild robot manipulation dataset.
\newblock \emph{arXiv preprint arXiv:2403.12945}, 2024.
\newblock URL \url{https://arxiv.org/abs/2403.12945}.

\bibitem[Yang et~al.(2026{\natexlab{a}})Yang, Li, Sun, Miao, Yang, Luo, Yan,
  Jiang, Shi, Fu, Chen, Zhao, Wang, Yao, Zhang, Chen, Li, and Zhu]{aoe2026}
Bowen Yang, Zishuo Li, Yang Sun, Changtao Miao, Yifan Yang, Man Luo, Xiaotong
  Yan, Feng Jiang, Jinchuan Shi, Yankai Fu, Ning Chen, Junkai Zhao, Pengwei
  Wang, Guocai Yao, Shanghang Zhang, Hao Chen, Zhe Li, and Kai Zhu.
\newblock {AoE}: Always-on egocentric human video collection for embodied {AI}.
\newblock In \emph{Proceedings of the IEEE/CVF Conference on Computer Vision
  and Pattern Recognition Workshops}, pages 3893--3902, 2026{\natexlab{a}}.

\bibitem[Liu et~al.(2022)Liu, Liu, Jiang, Lyu, Wan, Shen, Liang, Fu, Wang, and
  Yi]{liu2022hoi4d}
Yunze Liu, Yun Liu, Che Jiang, Kangbo Lyu, Weikang Wan, Hao Shen, Boqiang
  Liang, Zhoujie Fu, He~Wang, and Li~Yi.
\newblock {HOI4D}: A 4d egocentric dataset for category-level human-object
  interaction.
\newblock In \emph{Proceedings of the IEEE/CVF Conference on Computer Vision
  and Pattern Recognition}, pages 21013--21022, 2022.
\newblock URL \url{https://arxiv.org/abs/2203.01577}.

\bibitem[Banerjee et~al.(2025)Banerjee, Shkodrani, Moulon, Hampali, Han, Zhang,
  Zhang, Fountain, Miller, Basol, Newcombe, Wang, Engel, and
  Hodan]{banerjee2025hot3d}
Prithviraj Banerjee, Sindi Shkodrani, Pierre Moulon, Shreyas Hampali, Shangchen
  Han, Fan Zhang, Linguang Zhang, Jade Fountain, Edward Miller, Selen Basol,
  Richard Newcombe, Robert Wang, Jakob~Julian Engel, and Tomas Hodan.
\newblock {HOT3D}: Hand and object tracking in 3d from egocentric multi-view
  videos.
\newblock In \emph{Proceedings of the IEEE/CVF Conference on Computer Vision
  and Pattern Recognition}, pages 7061--7071, 2025.
\newblock URL \url{https://arxiv.org/abs/2411.19167}.

\bibitem[Kareer et~al.(2024)Kareer, Patel, Punamiya, Mathur, Cheng, Wang,
  Hoffman, and Xu]{kareer2024egomimic}
Simar Kareer, Dhruv Patel, Ryan Punamiya, Pranay Mathur, Shuo Cheng, Chen Wang,
  Judy Hoffman, and Danfei Xu.
\newblock {EgoMimic}: Scaling imitation learning via egocentric video.
\newblock \emph{arXiv preprint arXiv:2410.24221}, 2024.
\newblock URL \url{https://arxiv.org/abs/2410.24221}.

\bibitem[Zhang et~al.(2025)Zhang, Deng, Ma, and Potamias]{hawor2025}
Jinglei Zhang, Jiankang Deng, Chao Ma, and Rolandos~Alexandros Potamias.
\newblock {HaWoR}: World-space hand motion reconstruction from egocentric
  videos.
\newblock \emph{arXiv preprint arXiv:2501.02973}, 2025.
\newblock URL \url{https://arxiv.org/abs/2501.02973}.

\bibitem[Wang et~al.(2026)Wang, Ren, Morgan, Chen, Qian, Chanrungmaneekul, and
  Hang]{egoinfinity2026}
Gaotian Wang, Kejia Ren, Andrew Morgan, Yiting Chen, Howard~H. Qian, Podshara
  Chanrungmaneekul, and Kaiyu Hang.
\newblock {EgoInfinity}: A web-scale 4d hand-object interaction data engine for
  any-view robot retargeting and video-to-action robot learning.
\newblock \emph{arXiv preprint arXiv:2606.17385}, 2026.
\newblock URL \url{https://arxiv.org/abs/2606.17385}.

\bibitem[Niu et~al.(2026)Niu, Lv, Zhang, Wan, Gao, Ai, Xu, Hu, Zhang, Xie,
  Zhaxizhuoma, Zhao, Bing, Ding, and Liu]{niu2026egoaero}
Yichen Niu, Haoran Lv, Xinrui Zhang, Xueyao Wan, Shiyu Gao, Ying Ai, Hui Xu,
  Yongqi Hu, Hengyi Zhang, Yang Xie, Zhaxizhuoma, Yue Zhao, Zhenshan Bing, Yan
  Ding, and Jianxing Liu.
\newblock {EgoAERO}: Learning dexterous manipulation from a single egocentric
  video without object assets.
\newblock \emph{arXiv preprint arXiv:2606.08057}, 2026.
\newblock URL \url{https://arxiv.org/abs/2606.08057}.

\bibitem[Liu et~al.(2026)Liu, Cheng, Yin, Shin, Cueva, Yang, Chen, Zhang, and
  Xu]{liu2026egoengine}
Yangcen Liu, Shuo Cheng, Xinchen Yin, Woo~Chul Shin, Alfred Cueva, Yiran Yang,
  Zhenyang Chen, Chuye Zhang, and Danfei Xu.
\newblock {EgoEngine}: From egocentric human videos to high-fidelity dexterous
  robot demonstrations.
\newblock \emph{arXiv preprint arXiv:2606.12604}, 2026.
\newblock URL \url{https://arxiv.org/abs/2606.12604}.

\bibitem[Lepert et~al.(2025)Lepert, Fang, and Bohg]{phantom2025}
Marion Lepert, Jiaying Fang, and Jeannette Bohg.
\newblock Phantom: Training robots without robots using only human videos.
\newblock \emph{arXiv preprint arXiv:2503.00779}, 2025.
\newblock URL \url{https://arxiv.org/abs/2503.00779}.

\bibitem[Luo et~al.(2026)Luo, Wang, Zhang, Zheng, Xi, Xu, Xu, Yuan, Zhang,
  Wang, Feng, and Lu]{luo2026beingh05}
Hao Luo, Ye~Wang, Wanpeng Zhang, Sipeng Zheng, Ziheng Xi, Chaoyi Xu, Haiweng
  Xu, Haoqi Yuan, Chi Zhang, Yiqing Wang, Yicheng Feng, and Zongqing Lu.
\newblock {Being-H0.5}: Scaling human-centric robot learning for
  cross-embodiment generalization.
\newblock \emph{arXiv preprint arXiv:2601.12993}, 2026.
\newblock URL \url{https://arxiv.org/abs/2601.12993}.

\bibitem[Yu et~al.(2025)Yu, Shentu, Wu, Abbeel, Goldberg, and Wu]{yu2025egomi}
Justin Yu, Yide Shentu, Di~Wu, Pieter Abbeel, Ken Goldberg, and Philipp Wu.
\newblock {EgoMI}: Learning active vision and whole-body manipulation from
  egocentric human demonstrations.
\newblock \emph{arXiv preprint arXiv:2511.00153}, 2025.
\newblock URL \url{https://arxiv.org/abs/2511.00153}.

\bibitem[Yang et~al.(2026{\natexlab{b}})Yang, Bao, Xin, Song, Tian, Zhao, Wang,
  and Li]{yang2026zerowbc}
Haoran Yang, Jiacheng Bao, Yucheng Xin, Haoming Song, Yuyang Tian, Bin Zhao,
  Dong Wang, and Xuelong Li.
\newblock {ZeroWBC}: Learning natural whole-body humanoid interaction from
  human egocentric data.
\newblock \emph{arXiv preprint arXiv:2603.09170}, 2026{\natexlab{b}}.
\newblock URL \url{https://arxiv.org/abs/2603.09170}.

\bibitem[Shi et~al.(2026)Shi, Peng, Chen, Jiang, Li, Huang, Luo, Li, and
  Chen]{shi2026egohumanoid}
Modi Shi, Shijia Peng, Jin Chen, Haoran Jiang, Tianyu Li, Di~Huang, Ping Luo,
  Hongyang Li, and Li~Chen.
\newblock {EgoHumanoid}: Unlocking in-the-wild loco-manipulation with
  robot-free egocentric demonstration.
\newblock \emph{arXiv preprint arXiv:2602.10106}, 2026.
\newblock URL \url{https://arxiv.org/abs/2602.10106}.

\bibitem[Luo et~al.(2025)Luo, Feng, Zhang, Zheng, Wang, Yuan, Liu, Xu, Jin, and
  Lu]{luo2025being}
Hao Luo, Yicheng Feng, Wanpeng Zhang, Sipeng Zheng, Ye~Wang, Haoqi Yuan,
  Jiazheng Liu, Chaoyi Xu, Qin Jin, and Zongqing Lu.
\newblock {Being-H0}: Vision-language-action pretraining from large-scale human
  videos.
\newblock \emph{arXiv preprint arXiv:2507.15597}, 2025.
\newblock URL \url{https://arxiv.org/abs/2507.15597}.

\bibitem[Yang et~al.(2025)Yang, Yu, Wu, Yan, Li, Cheng, Zou, Fang, Cheng, Qiu,
  Yin, Liu, Han, Lu, and Wang]{egovla2025}
Ruihan Yang, Qinxi Yu, Yecheng Wu, Rui Yan, Borui Li, An-Chieh Cheng, Xueyan
  Zou, Yunhao Fang, Xuxin Cheng, Ri-Zhao Qiu, Hongxu Yin, Sifei Liu, Song Han,
  Yao Lu, and Xiaolong Wang.
\newblock {EgoVLA}: Learning vision-language-action models from egocentric
  human videos.
\newblock \emph{arXiv preprint arXiv:2507.12440}, 2025.
\newblock URL \url{https://arxiv.org/abs/2507.12440}.

\bibitem[Li et~al.(2025)Li, Deng, Liang, Luo, Zhou, Yao, Zeng, Feng, Liang, Xu,
  Zhang, Chen, Chen, Sun, Chen, Yang, and Guo]{vitra2025}
Qixiu Li, Yu~Deng, Yaobo Liang, Lin Luo, Lei Zhou, Chengtang Yao, Lingqi Zeng,
  Zhiyuan Feng, Huizhi Liang, Sicheng Xu, Yizhong Zhang, Xi~Chen, Hao Chen,
  Lily Sun, Dong Chen, Jiaolong Yang, and Baining Guo.
\newblock Scalable vision-language-action model pretraining for robotic
  manipulation with real-life human activity videos.
\newblock \emph{arXiv preprint arXiv:2510.21571}, 2025.
\newblock URL \url{https://arxiv.org/abs/2510.21571}.

\bibitem[Nikulin et~al.(2025)Nikulin, Zisman, Tarasov, Lyubaykin, Polubarov,
  Kiselev, and Kurenkov]{laom2025}
Alexander Nikulin, Ilya Zisman, Denis Tarasov, Nikita Lyubaykin, Andrei
  Polubarov, Igor Kiselev, and Vladislav Kurenkov.
\newblock Latent action learning requires supervision in the presence of
  distractors.
\newblock \emph{arXiv preprint arXiv:2502.00379}, 2025.
\newblock URL \url{https://arxiv.org/abs/2502.00379}.

\bibitem[Chen et~al.(2026)Chen, Wang, Chen, Chen, Gao, Tang, Li, Liu, Yao, Li,
  Xu, and Yu]{chen2026lawam}
Jialei Chen, Kai Wang, Kang Chen, Shuaihang Chen, Feng Gao, Wenhao Tang,
  Zhiyuan Li, Weilin Liu, Zhuyu Yao, Boxun Li, Yuanbo Xu, and Chao Yu.
\newblock {LaWAM}: Latent world action models for efficient dynamics-aware
  robot policies.
\newblock \emph{arXiv preprint arXiv:2606.15768}, 2026.
\newblock URL \url{https://arxiv.org/abs/2606.15768}.

\bibitem[Ye et~al.(2026)Ye, Ge, Zheng, Gao, Yu, Kurian, Indupuru, Tan, Zhu,
  Xiang, Malik, Lee, Liang, Ranawaka, Gu, Xu, Wang, Hu, Narayan, Bj{\"o}rck,
  Wang, Kim, Niu, Zheng, Xie, Wu, Wang, Julian, Xu, Du, Chebotar, Reed, Kautz,
  Zhu, Fan, and Jang]{dreamzero2026}
Seonghyeon Ye, Yunhao Ge, Kaiyuan Zheng, Shenyuan Gao, Sihyun Yu, George
  Kurian, Suneel Indupuru, You~Liang Tan, Chuning Zhu, Jiannan Xiang, Ayaan
  Malik, Kyungmin Lee, William Liang, Nadun Ranawaka, Jiasheng Gu, Yinzhen Xu,
  Guanzhi Wang, Fengyuan Hu, Avnish Narayan, Johan Bj{\"o}rck, Jing Wang,
  Gwanghyun Kim, Dantong Niu, Ruijie Zheng, Yuqi Xie, Jimmy Wu, Qi~Wang, Ryan
  Julian, Danfei Xu, Yilun Du, Yevgen Chebotar, Scott Reed, Jan Kautz, Yuke
  Zhu, Linxi Fan, and Joel Jang.
\newblock World action models are zero-shot policies.
\newblock \emph{arXiv preprint arXiv:2602.15922}, 2026.
\newblock URL \url{https://arxiv.org/abs/2602.15922}.

\bibitem[Gao et~al.(2026)Gao, Liang, Zheng, Malik, Ye, Yu, Tseng, Dong, Mo,
  Lin, Ma, Nah, Magne, Xiang, Xie, Zheng, Niu, Tan, Zentner, Kurian, Indupuru,
  Jannaty, Gu, Zhang, Malik, Abbeel, Liu, Zhu, Jang, and Fan]{dreamdojo2026}
Shenyuan Gao, William Liang, Kaiyuan Zheng, Ayaan Malik, Seonghyeon Ye, Sihyun
  Yu, Wei-Cheng Tseng, Yuzhu Dong, Kaichun Mo, Chen-Hsuan Lin, Qianli Ma,
  Seungjun Nah, Loic Magne, Jiannan Xiang, Yuqi Xie, Ruijie Zheng, Dantong Niu,
  You~Liang Tan, K.~R. Zentner, George Kurian, Suneel Indupuru, Pooya Jannaty,
  Jinwei Gu, Jun Zhang, Jitendra Malik, Pieter Abbeel, Ming-Yu Liu, Yuke Zhu,
  Joel Jang, and Linxi Fan.
\newblock {DreamDojo}: A generalist robot world model from large-scale human
  videos.
\newblock \emph{arXiv preprint arXiv:2602.06949}, 2026.
\newblock URL \url{https://arxiv.org/abs/2602.06949}.

\bibitem[Wu et~al.(2024)Wu, Yin, Feng, He, Li, Hao, and Long]{ivideogpt2024}
Jialong Wu, Shaofeng Yin, Ningya Feng, Xu~He, Dong Li, Jianye Hao, and
  Mingsheng Long.
\newblock {iVideoGPT}: Interactive videogpts are scalable world models.
\newblock \emph{arXiv preprint arXiv:2405.15223}, 2024.
\newblock URL \url{https://arxiv.org/abs/2405.15223}.

\bibitem[Guo et~al.(2025)Guo, Shi, Chen, and Finn]{ctrlworld2025}
Yanjiang Guo, Lucy~Xiaoyang Shi, Jianyu Chen, and Chelsea Finn.
\newblock {Ctrl-World}: A controllable generative world model for robot
  manipulation.
\newblock \emph{arXiv preprint arXiv:2510.10125}, 2025.
\newblock URL \url{https://arxiv.org/abs/2510.10125}.

\bibitem[Li et~al.(2026{\natexlab{b}})Li, Zhu, Pollefeys, and
  Barath]{droidw2026}
Moyang Li, Zihan Zhu, Marc Pollefeys, and Daniel Barath.
\newblock {DROID-SLAM} in the wild.
\newblock In \emph{Proceedings of the IEEE/CVF Conference on Computer Vision
  and Pattern Recognition}, 2026{\natexlab{b}}.
\newblock URL \url{https://arxiv.org/abs/2603.19076}.

\bibitem[Romero et~al.(2017)Romero, Tzionas, and Black]{mano2017}
Javier Romero, Dimitrios Tzionas, and Michael~J. Black.
\newblock Embodied hands: Modeling and capturing hands and bodies together.
\newblock \emph{ACM Transactions on Graphics}, 36\penalty0 (6), 2017.
\newblock \doi{10.1145/3130800.3130883}.

\bibitem[Paliwal et~al.(2026)Paliwal, Etukuru, Liang, Abbeel, Shafiullah, and
  Malik]{doasido2026}
Bhawna Paliwal, Haritheja Etukuru, William Liang, Pieter Abbeel, Nur
  Muhammad~Mahi Shafiullah, and Jitendra Malik.
\newblock Do as i do: Dexterous manipulation data from everyday human videos.
\newblock \emph{arXiv preprint arXiv:2606.19333}, 2026.
\newblock URL \url{https://arxiv.org/abs/2606.19333}.

\bibitem[Qin et~al.(2023)Qin, Yang, Huang, Van~Wyk, Su, Wang, Chao, and
  Fox]{dexretarget}
Yuzhe Qin, Wei Yang, Binghao Huang, Karl Van~Wyk, Hao Su, Xiaolong Wang, Yu-Wei
  Chao, and Dieter Fox.
\newblock {AnyTeleop}: A general vision-based dexterous robot arm-hand
  teleoperation system.
\newblock In \emph{Robotics: Science and Systems}, 2023.
\newblock URL \url{https://github.com/dexsuite/dex-retargeting}.

\bibitem[Pan et~al.(2025)Pan, Wang, Qi, Liu, Bharadhwaj, Sharma, Wu, Shi,
  Malik, and Hogan]{spider2025}
Chaoyi Pan, Changhao Wang, Haozhi Qi, Zixi Liu, Homanga Bharadhwaj, Akash
  Sharma, Tingfan Wu, Guanya Shi, Jitendra Malik, and Francois Hogan.
\newblock {SPIDER}: Scalable physics-informed dexterous retargeting.
\newblock \emph{arXiv preprint arXiv:2511.09484}, 2025.
\newblock URL \url{https://arxiv.org/abs/2511.09484}.

\bibitem[Cadene et~al.(2026)Cadene, Aliberts, Capuano, Aractingi, Zouitine,
  Kooijmans, Choghari, Russi, Pascal, Palma, Shukor, Moss, Soare, Aubakirova,
  Lhoest, Gallouedec, and Wolf]{cadene2026lerobot}
Remi Cadene, Simon Aliberts, Francesco Capuano, Michel Aractingi, Adil
  Zouitine, Pepijn Kooijmans, Jade Choghari, Martino Russi, Caroline Pascal,
  Steven Palma, Mustafa Shukor, Jess Moss, Alexander Soare, Dana Aubakirova,
  Quentin Lhoest, Quentin Gallouedec, and Thomas Wolf.
\newblock {LeRobot}: An open-source library for end-to-end robot learning.
\newblock \emph{arXiv preprint arXiv:2602.22818}, 2026.
\newblock URL \url{https://arxiv.org/abs/2602.22818}.

\bibitem[Bi et~al.(2025)Bi, Wu, Lin, Tan, Su, Su, and Zhu]{hrdt2025}
Hongzhe Bi, Lingxuan Wu, Tianwei Lin, Hengkai Tan, Zhizhong Su, Hang Su, and
  Jun Zhu.
\newblock {H-RDT}: Human manipulation enhanced bimanual robotic manipulation.
\newblock \emph{arXiv preprint arXiv:2507.23523}, 2025.
\newblock URL \url{https://arxiv.org/abs/2507.23523}.

\bibitem[Llontop and Neel(2026)]{grootn172026}
Edith Llontop and Brandon Neel.
\newblock Develop humanoid robot policies end-to-end with {NVIDIA Isaac GR00T}.
\newblock {NVIDIA} Technical Blog, July 2026.
\newblock URL
  \url{https://developer.nvidia.com/blog/develop-humanoid-robot-policies-end-to-end-with-nvidia-isaac-gr00t/}.
\newblock Official {GR00T N1.7} workflow and model overview.

\bibitem[Shukor et~al.(2025)Shukor, Aubakirova, Capuano, Kooijmans, Palma,
  Zouitine, Aractingi, Pascal, Russi, Marafioti, Alibert, Cord, Wolf, and
  Cadene]{smolvla2025}
Mustafa Shukor, Dana Aubakirova, Francesco Capuano, Pepijn Kooijmans, Steven
  Palma, Adil Zouitine, Michel Aractingi, Caroline Pascal, Martino Russi,
  Andres Marafioti, Simon Alibert, Matthieu Cord, Thomas Wolf, and Remi Cadene.
\newblock {SmolVLA}: A vision-language-action model for affordable and
  efficient robotics.
\newblock \emph{arXiv preprint arXiv:2506.01844}, 2025.
\newblock URL \url{https://arxiv.org/abs/2506.01844}.

\bibitem[Kazemi et~al.(2024)Kazemi, Savov, Paudel, and
  Van~Gool]{genieredux2024}
Naser Kazemi, Nedko Savov, Danda Paudel, and Luc Van~Gool.
\newblock Learning generative interactive environments by trained agent
  exploration.
\newblock \emph{arXiv preprint arXiv:2409.06445}, 2024.
\newblock URL \url{https://arxiv.org/abs/2409.06445}.

\bibitem[Li et~al.(2026{\natexlab{c}})Li, Zhang, Luo, Yang, Wang, Han, Yu, Gao,
  Xue, Zhu, Shen, and Xu]{lingbova2026}
Lin Li, Qihang Zhang, Yiming Luo, Shuai Yang, Ruilin Wang, Fei Han, Mingrui Yu,
  Zelin Gao, Nan Xue, Xing Zhu, Yujun Shen, and Yinghao Xu.
\newblock Causal world modeling for robot control.
\newblock \emph{arXiv preprint arXiv:2601.21998}, 2026{\natexlab{c}}.
\newblock URL \url{https://arxiv.org/abs/2601.21998}.

\bibitem[Gao et~al.(2025)Gao, Zhou, Du, Zhang, and Gan]{adaworld2025}
Shenyuan Gao, Siyuan Zhou, Yilun Du, Jun Zhang, and Chuang Gan.
\newblock {AdaWorld}: Learning adaptable world models with latent actions.
\newblock In \emph{Proceedings of the 42nd International Conference on Machine
  Learning}, volume 267 of \emph{Proceedings of Machine Learning Research},
  pages 18744--18771. PMLR, 2025.
\newblock URL \url{https://proceedings.mlr.press/v267/gao25u.html}.

\bibitem[Yoon et~al.(2026)Yoon, Han, Ji, Park, Kim, Kwon, and
  Kim]{egoxtreme2026}
Taegyoon Yoon, Yegyu Han, Seojin Ji, Jaewoo Park, Sojeong Kim, Taein Kwon, and
  Hyung-Sin Kim.
\newblock {EgoXtreme}: A dataset for robust object pose estimation in
  egocentric views under extreme conditions.
\newblock \emph{arXiv preprint arXiv:2603.25135}, 2026.
\newblock URL \url{https://arxiv.org/abs/2603.25135}.

\bibitem[Radford et~al.(2021)Radford, Kim, Hallacy, Ramesh, Goh, Agarwal,
  Sastry, Askell, Mishkin, Clark, Krueger, and Sutskever]{radford2021clip}
Alec Radford, Jong~Wook Kim, Chris Hallacy, Aditya Ramesh, Gabriel Goh,
  Sandhini Agarwal, Girish Sastry, Amanda Askell, Pamela Mishkin, Jack Clark,
  Gretchen Krueger, and Ilya Sutskever.
\newblock Learning transferable visual models from natural language
  supervision.
\newblock In \emph{Proceedings of the 38th International Conference on Machine
  Learning}, volume 139 of \emph{Proceedings of Machine Learning Research},
  pages 8748--8763. PMLR, 2021.
\newblock URL \url{https://proceedings.mlr.press/v139/radford21a.html}.

\bibitem[Lauren{\c{c}}on et~al.(2024)Lauren{\c{c}}on, Tronchon, Cord, and
  Sanh]{laurencon2024idefics2}
Hugo Lauren{\c{c}}on, L{\'e}o Tronchon, Matthieu Cord, and Victor Sanh.
\newblock What matters when building vision-language models?
\newblock In \emph{Advances in Neural Information Processing Systems},
  volume~37, 2024.
\newblock \doi{10.52202/079017-2789}.
\newblock URL
  \url{https://proceedings.neurips.cc/paper_files/paper/2024/hash/a03037317560b8c5f2fb4b6466d4c439-Abstract-Conference.html}.

\end{thebibliography}

\newpage
\section*{Appendix}
\subsection*{Privacy and Ethical Statement}
\label{sec:privacy_ethics}

The collection and release of the Open-AoE dataset were conducted with explicit authorization from all data contributors. Before participating, each contributor was informed of the purpose of data collection, the types of data involved, the intended scientific use of the collected data, and its subsequent public release as part of the Open-AoE dataset. All contributors voluntarily consented to the collection, processing, research use, and open-source distribution of their contributed data.

Before uploading, contributors reviewed the collected data locally on their mobile devices and explicitly confirmed each upload. After the confirmed data were uploaded, privacy-preserving processing was performed in a controlled cloud environment. This processing included masking visual regions that might reveal personal or sensitive information and removing or desensitizing metadata unrelated to embodied-intelligence research. The released data do not contain account information, payment information, or other personal metadata. Only privacy-processed, anonymized, and task-relevant data are made publicly available.

Open-AoE is released exclusively to support scientific research and technological development in embodied intelligence, robotics, and related fields. The dataset is not intended for identity recognition, personal profiling, surveillance, re-identification, or other uses that may infringe upon individual rights. Dataset users are expected to comply with the applicable license, privacy regulations, and generally accepted research-ethics principles, and must not attempt to recover or infer the identities of data contributors.

Based on the informed-consent process, explicit contributor authorization, local review and upload confirmation, controlled cloud-based privacy processing, data minimization, and anonymization measures described above, we have identified no unresolved material privacy or ethical concerns associated with the research use and open release of the Open-AoE dataset.

\end{document}